%% file: main.tex
\documentclass{article}
\usepackage{iclr2027_conference,times}
\usepackage{graphicx}
\usepackage{amsmath,amssymb}
\usepackage{booktabs,array,tabularx,longtable}
\usepackage{algorithm,algorithmic}
\usepackage{float}
\usepackage{url}
\usepackage{hyperref}
\newcolumntype{L}[1]{>{\raggedright\arraybackslash}p{#1}}
\newcolumntype{C}[1]{>{\centering\arraybackslash}p{#1}}
\hypersetup{pdftitle={NarraSpace: Multimodal Scene-Grounded Evidence for Picture-Description Dementia Classification},pdfauthor={Qingming Li, Ruiyan Sun, Satoshi Nakamura},pdfsubject={Research preprint},colorlinks=true,linkcolor=blue,citecolor=blue,urlcolor=blue}

\title{NarraSpace: Multimodal Scene-Grounded Evidence for Picture-Description Dementia Classification}
\iclrfinalcopy
\input{authors}
\begin{document}
\maketitle
\pagestyle{plain}
\thispagestyle{plain}
\begin{abstract}
Picture-description dementia classification uses a shared visual scene, but a transcript-level prediction alone does not identify the scene evidence underlying its inputs. NarraSpace links words and speech times to named picture elements and retrieves utterances for predefined scene events. Narrative-spatial descriptors summarize coverage and temporal organization. A frozen language model contrasts each event's text under examples from Alzheimer's disease and healthy-control training participants, alongside transcript surprisal and coverage measurements. Retained source records link event scores to their source utterances, speech intervals, and lexical matches. On the official 48-participant test set of the Alzheimer's Dementia Recognition through Spontaneous Speech benchmark, a fixed support vector classifier achieves $80.88\%\pm2.56\%$ mean accuracy over fifty seeds; averaged probabilities yield $81.25\%$ accuracy and $91.32\%$ area under the receiver operating characteristic curve. NarraSpace makes the construction of scene-linked classification inputs explicit and inspectable by connecting event-indexed language-model measurements to their source utterances, speech timing, and extraction records. Such source-linked measurements could support clinician review of picture descriptions by presenting quantitative summaries alongside the original speech evidence.
\end{abstract}

\section{Introduction}
\label{sec:introduction}

Picture-description dementia classification has a useful constraint: the picture supplies a shared reference for what a participant could describe. In the Alzheimer's Dementia Recognition through Spontaneous Speech (ADReSS) benchmark, every participant describes the Cookie Theft scene \citep{luzAlzheimersDementiaRecognition2020}. Descriptions differ in the objects mentioned, their order, the time spent describing them, and the relations expressed. Clinical picture-description assessment counts information units: depicted objects, actions, and facts conveyed by the speaker \citep{goodglass2001assessment,CROISILE19961,fraserLinguisticFeaturesIdentify2016}. These analyses suggest a representation question: how can a classifier's input retain the connection between a linguistic measurement and the particular scene content from which it was obtained?

Acoustic and linguistic features, pretrained representations, and whole-transcript language-model scores provide useful classification signals \citep{yuanPausesDetectionAlzheimers2021,martincTemporalIntegrationText2021,wangExploitingPromptLearning2023,xiaoAlzheimersDementiaDetection2025}. A scalar score, however, does not itself identify the phrase, event, or scene location associated with that signal. Prior work already represents picture descriptions as spatio-semantic paths, aligns spoken words with image regions, and models image--text consistency \citep{ambadiSpatioSemanticGraphsPicture2021a,heidarzadehEyeTrackingWordsAlzheimers,ngAutomatedExtractionSpatioSemantic2025a,liDetectingNeurocognitiveDisorders2025}. The connection we study is between an event in the picture, its retrieved utterances, and their likelihood under class-derived example contexts. Keeping these together allows the score and its extraction choices to be inspected.

NarraSpace constructs an evidence representation around a fixed scene schema. It links words to speech intervals and fifteen named picture elements (scene anchors), retrieves utterances for seven predefined events, and contrasts the same retrieved text under examples from Alzheimer's disease and healthy-control training participants using one frozen language model. A conventional classifier combines this evidence with speech, linguistic, and image--text descriptors. Retained source records allow a reader to move from a displayed event score back to its utterances, lexical matches, and scene anchors. This is \emph{source attribution}: it specifies where evidence came from. It is distinct from attributing a classifier's decision to individual features or demonstrating that an explanation improves clinical judgment.

Our contribution connects scene structure, event-indexed language-model measurements, and source records. We provide (1) a representation separating anchor trajectories, event retrieval, and classification features; (2) paired likelihood measurements for retrieved utterances, alongside scene-token and coverage descriptors; and (3) an assessment with shared transcript processing, family removals, participant-level uncertainty, and auditable cases. The event index lets a reader locate a measurement's source and inspect its extraction choices. Whether this organization improves inspection accuracy or speed over simple retrieval remains an empirical question.

\section{Related Work}
\label{sec:related_work}

\paragraph{Speech and language representations.}
ADReSS standardizes picture-description classification with an official participant split \citep{luzAlzheimersDementiaRecognition2020}. Relevant representations include linguistic and acoustic descriptors, temporal integration, pause-aware language models, ensembles, and prompt learning \citep{fraserLinguisticFeaturesIdentify2016,yuanPausesDetectionAlzheimers2021,syedAutomatedRecognitionAlzheimers2021,martincTemporalIntegrationText2021,wangExploitingPromptLearning2023}. Word-level multimodal alignment provides another route to relating the speech signal to its transcription \citep{ortiz-perezCogniAlignWordLevelMultimodal2025a}. Transcript construction matters: manual and automatic transcripts, annotation handling, and retained disfluencies can change the information available to a model \citep{liuManualTranscriptsPotential2025}. We use official participant transcripts and specify their normalization to make the input available to each feature family explicit.

\paragraph{Scene structure and contextual evidence.}
Spatio-semantic graphs encode paths through picture information units \citep{ambadiSpatioSemanticGraphsPicture2021a,ngAutomatedExtractionSpatioSemantic2025a}. A separate study by \citet{ngAdvancingSpatioSemantic2025} learns information-unit detection and ordering with a pretrained text encoder, evaluating both against human annotations and validating derived features externally. Word--region alignment, eye movements, image--text relevance, and topic evolution also connect descriptions to scenes \citep{heidarzadehEyeTrackingWordsAlzheimers,wangExploringComplementaryNature2024,zhuEvaluatingPictureDescription2023,liDetectingNeurocognitiveDisorders2025}. NarraSpace builds on this structural premise: its fixed event records index class-context likelihood measurements and their source utterances.

\paragraph{Language-model evidence and explanations.}
\citet{xiaoAlzheimersDementiaDetection2025} contrast whole-transcript perplexities from class-conditioned models. \citet{zhu2024adversarial} use training-derived feature context, including picture content, for adversarial generation and perplexity-based detection; \citet{su2026beyond} study demonstration-centered anchoring. NarraSpace applies class-derived contexts within one frozen model to event text and retains each contrast's source. The event, scored text, contrast, and extraction rules are available together for inspection. This measurement-level contribution complements the decision explanations studied in clinician-facing systems \citep{ntampakisNeuroXVocalDetectionExplanation2026a}.

\section{Method}
\label{sec:method}

\begin{figure}[t]
\centering
\includegraphics[width=\linewidth]{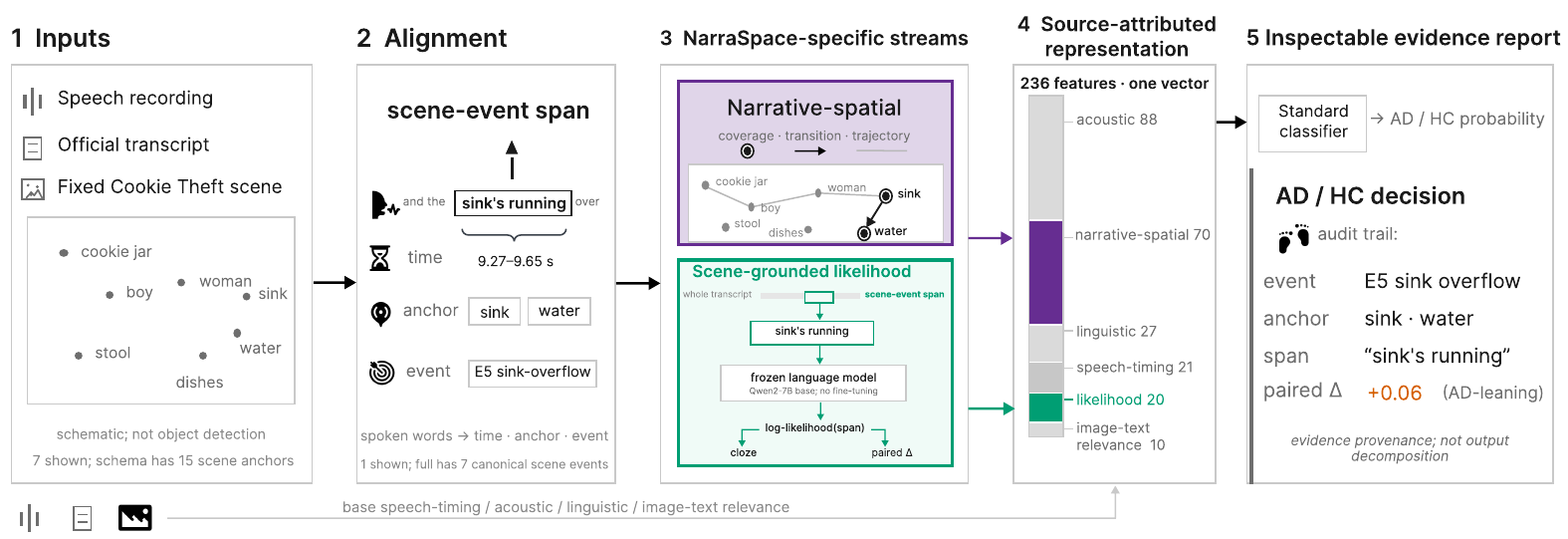}
\smallskip
{\small Classifier path: $236$ features $+\;20$ missingness flags
$\longrightarrow 256$ candidates $\longrightarrow 60$ selected inputs
$\longrightarrow$ classifier.}
\caption{NarraSpace constructs features while retaining links to their source text and speech times. The highlighted fragment illustrates a match; the language model scores the complete utterances retrieved for an event, concatenated in narrative order. The classifier path below the diagram expands its feature-selection step. Event-record summaries support source inspection and are excluded from classifier inputs (Appendix~\ref{sup:grounding_algorithm}). AD denotes Alzheimer's disease and HC healthy control.}
\label{fig:narraspace_overview}
\end{figure}

NarraSpace constructs classifier features and retains their source records
(Figure~\ref{fig:narraspace_overview}). A \emph{scene anchor} is a named
picture element, such as the sink. A \emph{visit} groups consecutive
mentions of one anchor along the timed word sequence. An \emph{event record}
groups utterances associated with a predefined scene event, such as sink
overflow; its five lexical fields record actor, action, object, location,
and consequence matches. These records remain available for inspection
even when their derived features are not selected by the classifier.

\subsection{Inputs, Scene Anchors, and Event Retrieval}
\label{subsec:alignment}

The inputs are audio, the official transcript, and the shared Cookie Theft
image. Transcript processing has two branches. Language-model scoring uses
participant speech after removing examiner turns, task instructions, annotation
markers, and angle-bracketed retraced text, including its words; retained
spoken forms are not replaced by standard-form corrections. Linguistic
features additionally use the original morphological and grammatical
annotations and raw disfluency markers. WhisperX forced alignment attaches
word times to the cleaned participant text without generating a new transcript
\citep{bainWhisperXTimeAccurateSpeech2023}.

Fifteen fixed scene anchors cover depicted entities, locations, and dish
drying (Appendix~\ref{sup:grounding_algorithm}). Words are assigned by phrase
rules, then lexical lookup, then text-embedding similarity. For the last
step, each anchor's \emph{prototype} averages embeddings of its natural-language
descriptions. The query is the target word with up to two preceding words
and one following word, or the word alone when that context has fewer than
two tokens. The \texttt{all-mpnet-base-v2} encoder supplies both representations.
Assignment requires cosine similarity at least $0.45$ and a top-two margin
at least $0.08$; other words remain unassigned. These text-to-schema matches assign the spoken content to fixed elements of
the shared picture.

Seven events cover cookie reaching, stool instability, girl observing,
dishwashing, overflow, unawareness of the hazard, and kitchen context.
An utterance is retrieved when any event-field lexicon or associated anchor
matches. Its complete text is retained. We call the concatenation of all
such utterances in narrative order the \emph{event text}; it may be
noncontiguous in the original transcript and shared across events. A field
is filled when its lexicon matches in any retrieved utterance. These lexical
matches neither resolve negation nor establish semantic entailment.

\subsection{Narrative-Spatial Measurements}
\label{subsec:narrative_spatial}

We compute seventy narrative-spatial features describing anchor coverage,
order, and timing. They form one part of the candidate pool: the full system
selects sixty inputs from the combined 256 candidates
(Section~\ref{subsec:diagnostic_representation}). Descriptions mentioning the
same anchors can differ in revisits or pauses. The inventory includes
coverage, path, graph, timing, mapping-quality summaries, and healthy-control
reference deviations; Appendix~\ref{sup:narrative_features} details its five
subfamilies.

After unassigned words are removed, assigned words are sorted by onset.
Adjacent entries with the same anchor form one visit, even across intervening
unassigned words or pauses; there is no gap threshold. Write visit $j$ as
$v_j=(r_j,t_j^{\mathrm{start}},t_j^{\mathrm{end}})$ and let $c(r)$ be the
anchor's fixed bounding-box center in reference-image pixels. Coverage is
$|\{r_j\}|/15$; dwell sums visit durations; transition distance is
$\|c(r_{j+1})-c(r_j)\|_2$. A revisit returns to an already visited anchor.
The resulting sequence describes how speech moves between scene anchors.

Three features compare coverage, first-mention order, and normalized dwell
allocation with a reference constructed from permitted healthy-control
training participants. They respectively use the fraction of reference anchors
not observed, one minus Kendall's rank correlation, and Kullback--Leibler
divergence. The reference is rebuilt within each evaluation partition.

\subsection{Scene-Grounded Likelihood}
\label{subsec:scene_grounded_likelihood}

\begin{figure}[t]
\centering
\includegraphics[width=0.85\linewidth]{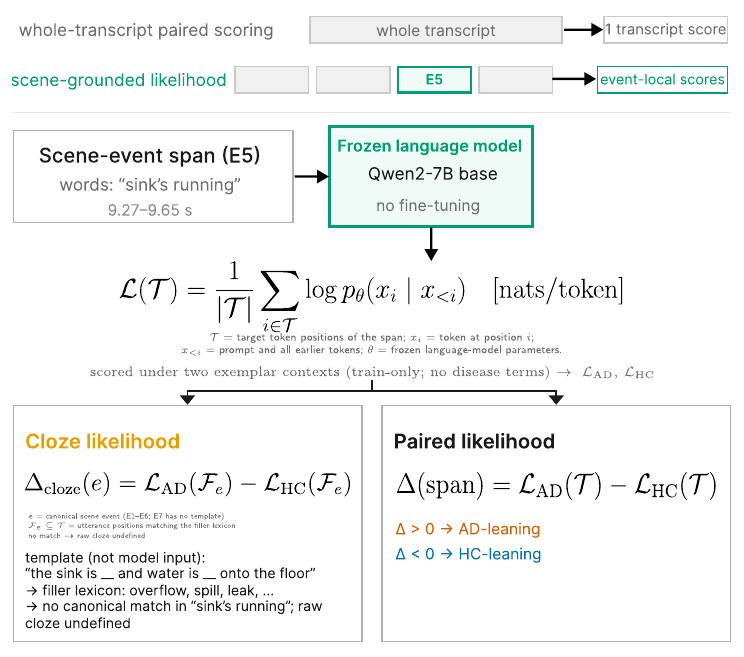}
\caption{Two inputs, one scoring model: AD training examples followed by the target event text, and HC training examples followed by that same text. Only target tokens are scored. Their mean log-likelihood difference gives an event contrast; the cloze contrast restricts the average to lexicon-matched subwords of that text. The illustrated excerpt is abridged, and the hand-written cloze template is not a model input.}
\label{fig:scene_grounded_likelihood}
\end{figure}

The twenty-feature likelihood stream combines event-text contrasts,
transcript-context surprisal, and lexical coverage. For the contrasts, a
frozen Qwen2-7B causal language model \citep{qwen2} scores the same event
text with examples from either Alzheimer's disease (AD) or healthy-control
(HC) training participants. Labels select the example pools but are absent
from the prompt; weights are not fine-tuned. This is supervised context
construction, illustrated in Figure~\ref{fig:scene_grounded_likelihood}.

For target-token positions $\mathcal{T}_e$ in event text $e$, the mean
log-likelihood under exemplar context $C$ is
\begin{equation}
\mathcal{L}_{C}(\mathcal{T}_e)
=\frac{1}{|\mathcal{T}_e|}\sum_{i\in\mathcal{T}_e}
\log p_{\theta}(x_i\mid C,x_{<i})
\quad [\mathrm{nats/token}].
\label{eq:span_likelihood}
\end{equation}
Tokenizer offsets separate target tokens from the unscored examples.
The paired event contrast is
\begin{equation}
\Delta_e=\mathcal{L}_{C_{\mathrm{AD}}}(\mathcal{T}_e)
-\mathcal{L}_{C_{\mathrm{HC}}}(\mathcal{T}_e).
\label{eq:paired_likelihood}
\end{equation}
Positive $\Delta_e$ means higher mean target log-likelihood under AD
examples; its scale is the log-likelihood difference in nats per token. For each event and
seed, two independently sampled AD/HC context pairs yield two contrasts,
which are averaged. Both classes use the same number of examples:
all $n$ if the smaller eligible pool has $n<10$, otherwise
$\min(30,\lfloor0.70n\rfloor)$, sampled without replacement.
Appendix~\ref{sup:likelihood} specifies the prompt and exclusions.

The \emph{cloze contrast} scores a lexicon-selected subset
$\mathcal{F}_e\subseteq\mathcal{T}_e$ using the same difference.
Hand-written templates supply the filler lexicons used to select these
positions, such as \emph{overflow}, \emph{spill}, and \emph{leak};
the selected positions are matching subword surfaces in the participant's
text, not necessarily all pieces of a matched word. Events 1--6 have these
lexicons; event 7 does not. A cloze score is undefined without matching
subwords; a paired score is undefined without retrieved utterances.

Four cloze summaries and nine paired descriptors contribute thirteen columns.
Six further columns summarize surprisal, $-\log p_{\theta}(x_i\mid x_{<i})$,
for scene-related words in their original transcript context; these use
whole-word intervals to select subwords. One unmatched-concept count completes
the twenty. The feature inventory in Appendix~\ref{sup:likelihood_features}
also separates the twelve score-derived columns from two coverage counts
and six surprisal summaries for the targeted controls. Removing the entire
stream therefore removes more than the paired-likelihood contrasts alone.

\subsection{Classifier Inputs and Probability Estimates}
\label{subsec:diagnostic_representation}

The representation has $146+70+20=236$ features. The 146 base features are
21 speech-timing, 88 acoustic, 27 linguistic, and 10 image--text descriptors.
Acoustics use the extended Geneva Minimalistic Acoustic Parameter Set with
openSMILE \citep{eybenGenevaMinimalisticAcoustic2016,eybenOpensmileMunichVersatile2010};
image--text features use frozen Contrastive Language--Image Pretraining
representations \citep{radfordLearningTransferableVisual2021}.
Three linguistic features also measure anchor coverage or repetition, so
removing narrative-spatial features does not remove all scene information.
Event-record summaries form a separate source-inspection branch
(Appendix~\ref{sup:grounding_algorithm}); they are excluded from this feature pool.

Each likelihood column is median-imputed using fitting rows and paired with
a binary missingness flag, giving $236+20=256$ candidates for the full system.
Selection combines two label-dependent rankings: larger mutual information
ranks first; smaller two-sided Mann--Whitney $p$-values rank first. The sixty
candidates with the lowest average ranks across the combined pool are
standardized and given to a support vector machine with radial basis
function kernel ($C=30$, $\gamma=0.001$). Family removals and targeted controls
also select sixty inputs from their respective pools; removing the likelihood
stream and its flags, for example, leaves 216 candidates.

The classifier internally fits a sigmoid to decision scores using five-fold
probability estimation. This internal operation does not repeat upstream
feature construction or selection; it differs from the outer validation
procedure below. Reports retain source records independently of selection; the selection
record identifies which derived features enter each classifier fit.

\section{Evaluation Setting}
\label{sec:experimental_setup}

\paragraph{Data and fixed system.}
ADReSS provides 108 official training and 48 official test participants.
All current analyses use the same fixed support vector machine ($C=30$,
$\gamma=0.001$, sixty selected inputs), including six whole-family removals.
Earlier evaluations fitted supervised feature selection outside validation
folds and incompletely excluded validation participants from example pools.
We corrected those fitting boundaries while retaining the classifier and
parameters; no replacement is selected from the new results. Earlier work
had already inspected the test set, so this remains an exploratory assessment.
Historical results are identified separately in Table~\ref{tab:main_comparison}B.

\paragraph{Training references and validation.}
Class-conditioned examples and the healthy-control reference depend on
training participants, so feature values depend on the evaluation partition.
For each of five fixed outer folds, we construct features for both fitting
and validation participants using only the other four folds. Every fitting
target is additionally excluded from its own examples and reference.
Imputation, selection, scaling, and classifier fitting use fitting rows only.
The held-out probabilities are concatenated to select thresholds.

For the final classifier, each training participant uses the feature version
constructed when its own fold was held out; all 108 such rows fit the final
preprocessing and classifier. Test participants instead use references from
all 108 training participants. These reference sizes differ. The classifier's
internal five-fold sigmoid fitting operates on the already constructed
features and does not refit upstream steps. Appendix~\ref{sup:evaluation}
tabulates these distinct scopes. Existing language-model scores are
reaggregated for the permitted pools; this correction performs no new
language-model inference.

\paragraph{Metrics and uncertainty.}
Fifty seeds vary example sampling and classifier randomness (classifier
states $237$--$286$); the participant fold partition stays fixed. Mean test
accuracy is primary, with population standard deviation across seeds.
Area under the receiver operating characteristic curve (AUC) measures ranking.
For each seed, the decision threshold maximizes Youden's statistic on the
training validation probabilities, using $0.10$--$0.90$ in $0.01$ steps and
choosing the smallest tied threshold \citep{Youden}.
\emph{Aggregated} predictions first average the fifty probabilities for each
participant, then apply a threshold selected from averaged training
probabilities. Threshold-optimized validation accuracy is not reported as
an unbiased estimate, since the same labels select the threshold.

Participant uncertainty uses 10,000 diagnosis-stratified bootstrap samples:
participants are sampled with replacement within each class, using identical
samples for both systems in a comparison. Fits and thresholds stay fixed.
For mean accuracy, each participant contributes its fraction of correct
seed-specific decisions; for aggregated accuracy, one correct/incorrect
decision; for aggregated AUC, its mean probability. The resulting 95\%
intervals concern performance or performance differences, not seed variation.
Exact two-sided McNemar tests use discordant aggregated decisions
\citep{McNemar_1947}. Neither analysis reverses prior test exposure; one test
participant still changes accuracy by $1/48$, or 2.08 percentage points.

\paragraph{Questions tested by feature removal.}
Whole-family removals test each family's contribution to the fitted system;
each refits selection and thresholds, and likelihood removal also deletes
its flags. Targeted controls ask whether event-score values add information
beyond coverage and surprisal, and whether compact or less redundant spatial
features suffice. They were specified after test exposure, retain all twenty
flags as candidates, and select no replacement primary system
(Appendix~\ref{sup:review_controls}). Historical whole-transcript and
permutation controls retain their earlier evaluation procedures.

\section{Results and Evidence Analysis}
\label{sec:results}

\subsection{Primary System and Reference Configurations}
\label{subsec:backend_comparison}

The current analyses ask how the representation classifies participants,
which feature groups affect those predictions, and what its source records
show. Table~\ref{tab:main_comparison}A presents the fixed primary system used
in the subsequent family-removal and participant-level analyses. Block B
shows three Xiao--Mistral reproductions and eight NarraSpace classifier
backends under their original evaluation procedures. Some NarraSpace
backends exceed the reconstructed Xiao--Mistral scores on individual
metrics; the largest aggregated accuracy in each group is $83.33\%$.
Different configurations attain the maxima across metrics.
Appendix~\ref{sup:restored_historical_comparison} details the fitting and
aggregation procedures; Appendix~\ref{sup:published_reference} defines the
three Xiao scores and lists original published results separately.

\begin{table}[t]
\caption{Classification performance on the 48 official-test participants. Block A presents the primary NarraSpace system; block B presents Xiao--Mistral reproductions and NarraSpace classifier backends under their original evaluation procedures. Values are percentages. Mean $\pm$ standard deviation summarizes fifty seeds; Best denotes the maximum for each metric. Bold marks column maxima within block B, including ties. Aggregation and evaluation procedures are specified below and in Appendix~\ref{sup:controls}.}
\label{tab:main_comparison}
\centering
\small
\begingroup
\setlength{\tabcolsep}{2.4pt}
\begin{tabular}{@{}L{0.25\linewidth}rrrrrr@{}}
\toprule
& \multicolumn{3}{c}{Accuracy (\%)} & \multicolumn{3}{c}{AUC (\%)} \\
\cmidrule(lr){2-4}\cmidrule(lr){5-7}
Method & Mean $\pm$ std. & Best & Aggregated & Mean $\pm$ std. & Best & Aggregated \\
\midrule
\multicolumn{7}{@{}l}{\textbf{A. Primary NarraSpace system}} \\
\midrule
Support vector machine (radial basis function)
& $80.88\pm2.56$ & 85.42 & 81.25 & $90.65\pm1.32$ & 93.23 & 91.32 \\
\midrule
\multicolumn{7}{@{}l}{\textbf{B. Reference configurations under their original evaluation procedures}} \\
\multicolumn{7}{@{}l}{\emph{Xiao--Mistral reproductions with harmonized transcripts}$^{\dagger}$} \\
\midrule
$\tilde{D}_{\log}$ & $80.92\pm4.40$ & 87.50 & 79.17 & $88.62\pm1.86$ & 93.23 & 89.41 \\
$\bar{D}$ & $80.67\pm4.27$ & 87.50 & \textbf{83.33} & $90.03\pm2.17$ & 94.27 & 91.84 \\
$\bar{D}^{*}$ & $80.50\pm4.48$ & 87.50 & \textbf{83.33} & $89.92\pm2.23$ & 94.27 & 87.33 \\
\midrule
\multicolumn{7}{@{}l}{\emph{NarraSpace classifier backends with the same 236-feature representation}$^{\ddagger}$} \\
\midrule
Support vector machine (radial basis function)
& $80.25\pm2.54$ & 85.42 & 81.25 & $90.70\pm1.30$ & 92.88 & 91.32 \\
Logistic regression & $81.29\pm1.93$ & 85.42 & 79.17 & $\mathbf{91.19}\pm0.90$ & 93.23 & 91.15 \\
Random forest & $78.67\pm2.69$ & 83.33 & 81.25 & $90.44\pm0.54$ & 92.01 & 90.45 \\
Extreme gradient boosting & $79.33\pm3.70$ & 87.50 & 81.25 & $87.38\pm2.17$ & 91.15 & 88.02 \\
Multilayer perceptron & $75.13\pm6.91$ & 87.50 & 81.25 & $84.59\pm6.88$ & \textbf{94.79} & \textbf{93.06} \\
Compact bottleneck classifier & $76.58\pm6.93$ & \textbf{91.67} & \textbf{83.33} & $83.51\pm6.69$ & 92.19 & 92.71 \\
Soft voting ensemble & $79.75\pm3.20$ & 85.42 & \textbf{83.33} & $87.23\pm1.10$ & 89.58 & 88.02 \\
Stacking ensemble & $\mathbf{81.46}\pm2.33$ & 85.42 & 81.25 & $90.37\pm1.27$ & 92.36 & 91.15 \\
\bottomrule
\end{tabular}
\endgroup

\smallskip
\begin{minipage}{\linewidth}
\small
$^{\dagger}$Xiao--Mistral averages perplexities before scoring and uses test-derived equal-error-rate thresholds. Definitions of $\tilde D_{\log}$, $\bar D$, and $\bar D^*$ are provided in Appendix~\ref{sup:published_reference}.
$^{\ddagger}$NarraSpace averages classifier probabilities and uses training-derived thresholds. Fitting procedures for blocks A and B are detailed in Appendices~\ref{sup:evaluation} and~\ref{sup:restored_historical_comparison}. AUC denotes area under the receiver operating characteristic curve.
\end{minipage}
\end{table}

\subsection{Classification and Evidence-Family Removals}
\label{subsec:classification_results}

Table~\ref{tab:corrected_main} reports the fixed classifier under the
partition-isolated procedure. The full representation achieves
$80.88\%\pm2.56\%$ mean accuracy and $90.65\%\pm1.32\%$ mean AUC.
Its aggregated decision rule correctly classifies 39 of 48 participants
($81.25\%$), with AUC $91.32\%$ and training-selected threshold $0.56$.
The participant-bootstrap 95\% interval is $[70.75,89.88]\%$ for mean
accuracy, $[68.75,91.67]\%$ for aggregated accuracy, and
$[82.47,97.57]\%$ for aggregated AUC. The wider bootstrap intervals reflect participant-sampling uncertainty
in the 48-person test set; the standard deviations describe variability
across construction seeds.

\begin{table}[t]
\caption{Corrected official-test results with the same support vector machine
and sixty selected inputs throughout. Values are percentages; $\pm$ is the
population standard deviation over fifty seeds. The full representation is
the primary system fixed before these family-removal comparisons.
Each removal refits preprocessing, feature selection, classifier, and
training-derived thresholds. AUC denotes area under the receiver operating
characteristic curve.}
\label{tab:corrected_main}
\centering
\small
\begin{tabular}{@{}lcccc@{}}
\toprule
 & \multicolumn{2}{c}{Fifty-seed mean} & \multicolumn{2}{c}{Aggregated probability} \\
\cmidrule(lr){2-3}\cmidrule(lr){4-5}
Removed evidence family & Accuracy & AUC & Accuracy & AUC \\
\midrule
None (full) & $80.88\pm2.56$ & $90.65\pm1.32$ & 81.25 & 91.32 \\
Speech timing & $83.25\pm2.24$ & $91.14\pm1.42$ & 85.42 & 92.19 \\
Acoustic & $81.08\pm2.49$ & $90.89\pm1.39$ & 79.17 & 91.15 \\
Linguistic & $80.88\pm2.63$ & $89.76\pm1.71$ & 81.25 & 90.97 \\
Image--text relevance & $81.25\pm2.57$ & $91.23\pm1.20$ & 79.17 & 91.84 \\
Narrative-spatial & $78.21\pm3.80$ & $86.80\pm1.69$ & 81.25 & 87.67 \\
Scene-grounded likelihood & $81.63\pm2.97$ & $88.89\pm0.02$ & 81.25 & 88.89 \\
\bottomrule
\end{tabular}
\end{table}

Removing the narrative-spatial family gives the largest decrease in mean
accuracy, from $80.88\%$ to $78.21\%$. The full-minus-removal difference is
$2.67$ percentage points, with paired 95\% interval $[-3.67,9.00]$; its
aggregated AUC difference is $3.65$ points, with interval $[-2.78,10.59]$.
Both intervals include zero, leaving the direction and magnitude of the
family's benefit uncertain. These estimates concern the seventy-feature
family as a whole. Aggregated accuracy is unchanged,
with three corrected and three regressed decisions.

Removing speech timing gives $83.25\%$ mean accuracy and 41 correct aggregated
decisions. Its $2.38$-point advantage has paired interval $[-0.46,6.63]$;
none of the six mean-accuracy intervals excludes zero. The specified full
system remains primary. Appendix~\ref{sup:corrected_statistics} gives all
paired intervals and decision counts for these alternative feature sets.

\paragraph{Targeted representation controls.}
We first ask whether event-score descriptors add information beyond coverage,
transcript surprisal, and missingness. The comparison retains all base and
narrative-spatial features, together with coverage and transcript-surprisal
descriptors and all missingness flags. Adding the twelve event-score
descriptors changes mean accuracy from $80.17\%$ to $80.88\%$, a difference
of $0.71$ percentage points.

A second control replaces the seventy narrative-spatial features with a
compact six-feature alternative, retaining all other feature groups and
missingness flags. Mean accuracy is $80.88\%$ for the full representation and
$77.96\%$ for the compact alternative, a difference of $2.92$ percentage
points; aggregated accuracy is $81.25\%$ and $83.33\%$, respectively.
For each candidate pool, feature selection, classifier fitting, and
training-derived threshold selection are repeated, with sixty selected
inputs throughout. These comparisons describe observed differences; paired
uncertainty estimates are reported in Appendix~\ref{sup:review_controls}.

\subsection{Participant-Level Decision Changes}
\label{subsec:likelihood_results}

Aggregate accuracy can conceal changes in which participants are classified
correctly. With and without the likelihood stream, the aggregated systems
each classify 39 of 48 participants correctly, yet their predictions differ
for ten participants. Adding the stream corrects five errors and introduces
five new errors. Four of the corrected decisions concern healthy controls,
whereas four of the new errors concern participants with Alzheimer's disease.
Thus, the unchanged aggregate accuracy accompanies a shift in the
distribution of errors across the two groups.

This comparison concerns the complete fitted systems, including their
training-selected thresholds. Table~\ref{tab:changed_likelihood_cases}
documents every changed decision and its associated probabilities;
Appendix~\ref{sup:corrected_statistics} reports the paired statistical
analysis. The event-score-specific comparison in
Section~\ref{subsec:classification_results} separately examines the
contribution of numerical event descriptors while retaining coverage and
missingness information.

\subsection{Inspecting Scene-Linked Measurements}

\begin{figure}[t]
\centering
\includegraphics[width=\linewidth]{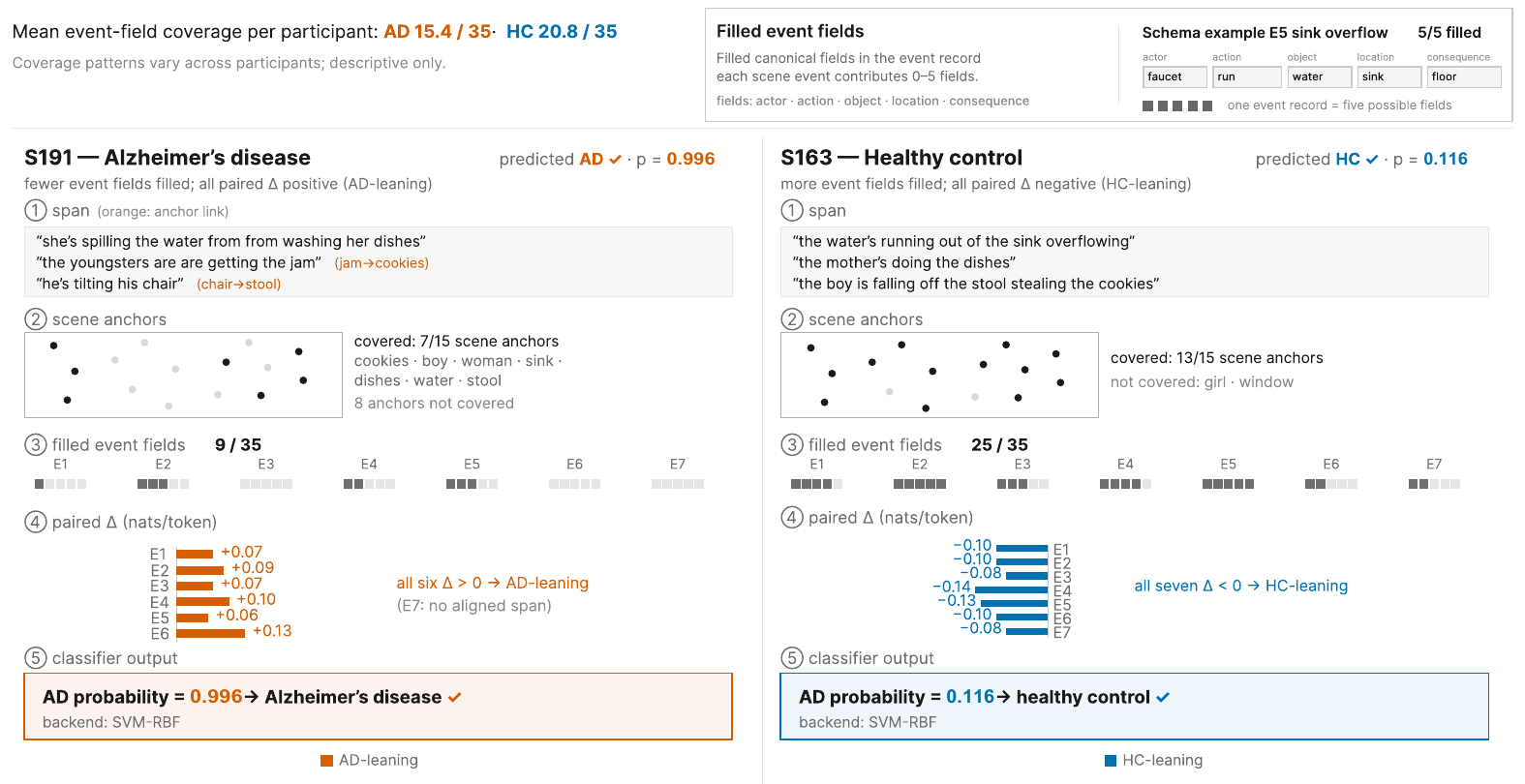}
\caption{Inspecting scene-linked measurements through their source speech. Reports for two test participants connect retrieved utterances with scene-anchor matches, event-field coverage, and likelihood contrasts. E1--E7 denote cookie reaching, stool instability, girl observing, dishwashing, overflow, unawareness, and kitchen context. Filled fields indicate lexical matches; excerpts are abridged, while scores use the full retrieved text. Header means summarize event-field coverage across the 48 official-test participants. For each participant, the displayed probability is the model-estimated probability of Alzheimer's disease, averaged across fifty runs. E7's ``no aligned span'' indicates an unavailable event-text contrast.}
\label{fig:traceable_cases}
\end{figure}

NarraSpace retains the link between a numerical measurement and the speech
from which it was constructed. Figure~\ref{fig:traceable_cases} illustrates
this connection through two scene events. For S163, the utterance ``the
water's running out of the sink overflowing'' is retrieved for sink overflow.
For S191, ``he's tilting his chair'' reaches the stool anchor through a
chair-to-stool match. Figure~\ref{fig:traceable_cases} presents these utterances alongside
scene-anchor matches, event-field coverage, and likelihood contrasts,
allowing readers to examine the textual evidence and mapping choices behind
the measurements.

These examples make different parts of feature construction accessible to
inspection: the overflow example exposes the text retrieved for an event,
while the chair example exposes the mapping between a participant's wording
and the predefined scene vocabulary. The feature-selection record identifies
which derived measurements enter a fitted classifier. Both participants are
classified correctly with and without the likelihood stream; the examples
illustrate how the inputs can be examined alongside the predictions.

A single-annotator audit of 404 accepted phrase-rule and prototype mappings
from 80 participants found 388 correct assignments, corresponding to
$96.04\%$ precision. Appendix~\ref{sup:mapping_audit} reports the sampling
procedure, mapping-tier results, and audit scope. Unfilled event fields
record that no expression matched the predefined lexical rules; the reports
therefore label them as \emph{undetected evidence}.

When an utterance is retrieved for multiple events, the records retain the
shared text so that its use across those events can be examined
(Appendix~\ref{sup:review_controls}). By retaining the original utterances
and their mapping records, NarraSpace makes the construction of scene-linked
classification inputs available for direct examination.

\section{Discussion}
\label{sec:limitations}

The shared picture provides a reference for connecting what a participant
says with how the description unfolds. NarraSpace uses scene anchors to
organize the timed word sequence and events to index retrieved utterances
and their class-context likelihood contrasts. Its contribution lies in
keeping these measurements connected to the speech and extraction choices
that produced them. A reader can examine the wording behind a score,
inspect a mapping between words and scene anchors, and identify text shared
by several events. These connections make the construction of classification
inputs explicit and available for review.

The empirical findings show why measurement construction and predictive
performance need to be examined together. Removing narrative-spatial features
lowers mean accuracy, while the compact alternative performs differently
across mean and aggregated accuracy. The likelihood stream changes which
participants are classified correctly even when aggregate accuracy is
unchanged. These observations motivate examining the selected features and
participant decisions alongside overall metrics; their paired uncertainty
estimates are reported in the appendices. Source records support this
examination at the level of input construction. Attributing a prediction to
a particular measurement additionally requires analysis of the fitted
classifier and its threshold.

The present evaluation is exploratory, using one English-language benchmark
with prior test exposure. The mapping audit assesses accepted anchor
assignments; direct event-level annotation and same-model comparisons of
local and whole-transcript scoring would test further aspects of the
representation. Independent participant cohorts would assess transfer beyond
this setting. For clinical research, source-linked measurements could help
reviewers examine quantitative summaries alongside the original speech,
including the wording and timing associated with a scene event. Evaluating
that use requires measuring inspection accuracy and effort, followed by
assessment of its contribution to clinical judgment.

\section{Conclusion}
\label{sec:conclusion}

NarraSpace connects scene structure, event-indexed language-model
measurements, and speech timing in a representation whose sources remain
available for inspection. On the official 48-participant test set, the fixed
classifier achieves $80.88\%$ mean accuracy over fifty seeds and $81.25\%$
accuracy after probability aggregation. Feature controls and participant-level
comparisons characterize how the representation affects classification,
while source-linked cases show how its measurements can be examined.
Together, the representation and evaluation provide a basis for studying
picture descriptions through both quantitative summaries and their original
speech evidence, with a concrete route toward evaluating their use in
human review.

\clearpage
\bibliography{references}
\bibliographystyle{iclr2027_conference}
\appendix
\section{Appendix}
\label{sup:impl}

This supplement specifies the feature construction, the scope of the mapping
audit, and the distinction between evidence provenance and classifier
explanation. The dataset is the Alzheimer's Dementia Recognition through
Spontaneous Speech (ADReSS) Cookie Theft benchmark, with 108 official training
participants and 48 official test participants
\citep{luzAlzheimersDementiaRecognition2020}. Alzheimer's disease (AD) is the
positive class and healthy control (HC) is the negative class. The reference
image, recordings, and transcripts are subject to their original access terms.

\subsection{Feature Accounting}
\label{sup:feature_accounting}

Table~\ref{tab:supp_feature_accounting} distinguishes constructed features,
selection candidates, and selected classifier inputs for the full system.
The representation contains $146+70+20=236$ features. Median imputation and
one missingness indicator for each of the twenty likelihood-stream columns
produce $236+20=256$ candidates. Feature selection retains
$K_{\mathrm{feat}}=60$ columns across this combined pool for the fixed support
vector machine with radial basis function kernel. The number selected from
each family depends on its ranks in that fit. Family removals and targeted
controls use their specified candidate pools and the same sixty-input selection.

\begin{table}[htbp]
\caption{Full-system feature accounting: 236 constructed features, 256 selection candidates, and sixty selected inputs. The 31 event-record summaries are separate from all three stages and are used for source inspection only.}
\label{tab:supp_feature_accounting}
\centering
\small
\begin{tabular}{@{}p{0.48\linewidth}rr@{}}
\toprule
Evidence family & Representation & Added indicators \\
\midrule
Speech timing & 21 & 0 \\
Acoustic & 88 & 0 \\
Linguistic & 27 & 0 \\
Image--text relevance & 10 & 0 \\
Narrative-spatial & 70 & 0 \\
Scene-likelihood and coverage & 20 & 20 \\
\midrule
Total & 236 & 20 \\
Classifier candidate pool & \multicolumn{2}{r}{256} \\
Selected classifier input & \multicolumn{2}{r}{60} \\
\bottomrule
\end{tabular}
\end{table}

The twenty-column stream contains four cloze summaries, nine paired-likelihood
features, six transcript-context surprisal summaries, and one unmatched-concept
count. This combines event scores with transcript and coverage measures. Removing
this stream removes all twenty semantic columns and their twenty missingness
indicators, leaving 216 base and narrative-spatial candidates. The stream
contributes twenty numeric columns and twenty flags to the full pool. A
separate set of 31 event-record summaries supports source inspection;
Appendix~\ref{sup:grounding_algorithm} gives their construction.

\subsection{Transcript Construction and Timing}
\label{sup:transcript}

Each official Codes for the Human Analysis of Transcripts (CHAT)
\texttt{.cha} file is reduced to participant-only words. Interviewer and
examiner turns and task instructions are excluded. The scored text removes
CHAT filler codes (\texttt{\&uh}, \texttt{\&um}), retracing and self-correction
markers (\texttt{[/]}, \texttt{[//]}), standard-form targets
(\texttt{[:\ ...]}), error tags (\texttt{[*\ ...]}), and unintelligibility
markers (\texttt{xxx}). It also removes complete angle-bracketed retraced
spans, including the words inside them. For example,
\texttt{<the boy> [/] the boy} becomes \texttt{the boy}; the first occurrence
is not retained. Spoken forms remaining after these rules are not replaced
with standard-form corrections. The procedure neither diagnoses paraphasias
nor preserves every original repetition.

Word timing uses WhisperX 3.8.5 in forced-alignment mode
\citep{bainWhisperXTimeAccurateSpeech2023}, with the participant transcript as
the target and the English \texttt{WAV2VEC2\_ASR\_BASE\_960H} checkpoint.
The timing stage does not generate the scored transcript using automatic
speech recognition. The anchor mapper and WhisperX produce separate word lists. To attach
WhisperX times to mapped words, their normalized spellings are compared using
Levenshtein similarity of at least $0.8$ and a length difference of at most two
characters. Multiword grounding phrases retain the interval from the first
aligned word to the last aligned word for their anchor token. The fallback divides an utterance interval among its remaining grounded
words when word matches are unavailable. The archived alignment check for
the 156 participants recorded no use of this fallback; the reported timings
therefore came from word matching, including phrase intervals. Such
timing estimates the allocation of speech to the mentioned scene anchors.

\subsection{Speech-Timing and Linguistic Base Features}
\label{sup:base_timing_linguistic}

The 21 speech-timing and 27 linguistic descriptors form 48 of the 146 base features. Demographic metadata are excluded from classifier inputs; pauses are represented by the timing statistics below.

\paragraph{Speech timing: twenty-one features.}
Nonnegative-timestamp words are processed in input order. Four descriptors are word count, first-onset-to-last-offset duration, and mean/population-standard-deviation word duration. Four rate descriptors are global words/second and local-rate mean, standard deviation, and coefficient of variation. Local rates count words intersecting inclusive five-second windows with a 2.5-second stride; only complete windows are used. Without one, the mean falls back to global rate and dispersion is zero. Ten pause descriptors summarize nonnegative consecutive inter-word gaps: mean, standard deviation, maximum, median, coefficient of variation, SciPy-default skewness and excess kurtosis, count and fraction exceeding one second, and count below 0.15 seconds. Zero gaps enter these statistics. Three further descriptors are summed word durations, summed gaps, and their pause-to-speech ratio. Empty word sequences yield zeros; skewness/kurtosis use zero when fewer than three gaps exist.

\paragraph{Linguistic evidence: twenty-seven features.}
Cleaned participant words supply lexical counts, but official CHAT morphological (\texttt{\%mor}) and grammatical (\texttt{\%gra}) annotations and raw utterance markers remain inputs to this family. Let $N$ denote words, $V$ lowercase types, and $V_1$ types occurring once.

Five lexical descriptors are $V/N$, its moving-window mean with window $\min(20,N)$ and unit stride, Brunet's $N^{V^{-0.172}}$, Honor\'e's $100\log(N)/(1-V_1/V)$, and $V_1/V$. Honor\'e's value defaults to zero when undefined. Eight morphological descriptors are noun, verb, adjective, and pronoun fractions; pronoun/noun, verb/noun, content-word, and light-verb ratios. The four part-of-speech fractions and content fraction use all parsed morphological tags; content comprises nouns, verbs, adjectives, and adverbs. Light verbs are \emph{go/get/do/make/take/have/be/come/put/give}, divided by verb count. With no nouns, numerator/noun ratios return one for a positive numerator, otherwise zero.

Five syntactic descriptors are mean utterance length, mean per-utterance dependency distance excluding root/punctuation arcs, maximum dependency depth, subordinate relations per utterance, and clausal relations per parsed dependency. Subordination counts \texttt{COMP,CJCT,CPRED,CMOD,XCOMP,XJCT,XMOD}; the clausal subset excludes the final two. Six disfluency descriptors are raw \texttt{\&word}, retracing (\texttt{[/]}), and revision (\texttt{[//]}) counts per word; incomplete (\texttt{+...}/\texttt{+/.}) and empty-utterance fractions; and filler/content-word ratio. Finally, three anchor-derived descriptors are the fraction of word records assigned an anchor (the denominator includes unassigned records), the number of distinct assigned anchors, and assigned word records divided by distinct anchors. An assigned word record is one anchor mention for these counts. Empty lexical input yields 27 zeros; absent mapping yields zero for its three descriptors. These annotation-dependent features require more than the normalized text scored by the language model.

\subsection{Acoustic Base Features}

The 88 acoustic descriptors are file-level functionals from openSMILE's
extended Geneva Minimalistic Acoustic Parameter Set, version 02
\citep{eybenGenevaMinimalisticAcoustic2016,eybenOpensmileMunichVersatile2010}.
They are computed once per benchmark-provided enhanced full recording.
This extractor does not trim the audio using the participant-only transcript
turns; transcript filtering and the acoustic recording boundary are distinct.

\subsection{Scene Anchors, Grounding, and Event Records}
\label{sup:grounding_algorithm}

The fixed schema contains fifteen anchors: boy, girl, woman, stool, cookie
jar, cookies, sink, water, dishes, curtain, window, counter, floor, garden,
and dish drying. The last anchor denotes an action; the other fourteen denote
depicted entities, elements, or locations. Each anchor therefore specifies
a category of scene content with a fixed image location. The schema operationalizes scene
content considered in picture-description and spatio-semantic analyses
\citep{CROISILE19961,fraserLinguisticFeaturesIdentify2016,
ambadiSpatioSemanticGraphsPicture2021a,ngAutomatedExtractionSpatioSemantic2025a}.

Eligible words are mapped in the order phrase rule, lexical lookup, then
prototype matching. Phrase rules take priority to preserve the assignment
of a complete multiword expression. Rule-specific anchor
positions determine which token receives the phrase assignment. The
prototype fallback uses \texttt{sentence-transformers/all-mpnet-base-v2}
through the Transformers tokenizer and model interfaces. Each anchor prototype is the normalized mean of normalized embeddings of
its natural-language descriptions. The query contains up to two words before
the target and one after it, excluding punctuation; when fewer than two
tokens remain, the target word alone is used. A candidate is accepted only if
its largest cosine similarity is at least $0.45$ and its margin over the
second-largest similarity is at least $0.08$. Tokens excluded as
non-groundable and eligible tokens that fail these gates remain distinguishable
in the mapping records. Neither state is a human-verified omission.

\begin{table}[t]
\caption{The seven canonical event records. Each has actor, action, object,
location, and consequence fields. Utterances are assigned through lexical
or anchor matches to this schema.}
\label{tab:supp_events}
\centering
\small
\begin{tabular}{@{}lp{0.77\linewidth}@{}}
\toprule
Identifier & Canonical scene event \\
\midrule
E1 & Boy reaching for cookies \\
E2 & Stool instability \\
E3 & Girl observing \\
E4 & Woman dishwashing \\
E5 & Sink overflow \\
E6 & Woman unaware of the hazard \\
E7 & Global kitchen context \\
\bottomrule
\end{tabular}
\end{table}

Event assignment is deterministic. An utterance supports an event when at
least one normalized token matches a term in any of that event's five field
lexicons, or when an accepted anchor overlaps the event's anchor set. The
entire supporting utterance is retained, and supporting utterances are
concatenated in narrative order to form the event text. A field is marked
present if its lexicon has a hit in any supporting utterance. The same
utterance can support several events; no cross-event deduplication is applied.
Retrieval retains complete utterances and accepts partial lexical evidence.
Actor--action relations and negation scope are outside the matching rule,
so a negated mention can still contribute a lexical or anchor hit.

For each participant, seven events with five fields give $7\times5=35$
event-field positions. These records yield $7\times3+5+5=31$ source-inspection
summaries. For each event, these include filled-field fraction, actor--action--object
co-occurrence, and absence of retrieved utterances
($21$). Co-occurrence means that all three field types have hits somewhere
in the event record, possibly in different utterances. Five descriptors
average each field type across events ($5$), followed by five overall
descriptors: mean completeness, missing consequence count, missing object
count, vague-term count, and event-order agreement. Their purpose is to
summarize recorded evidence for inspection. They are excluded from the 236
classification features, the 256 selection candidates, and the sixty selected inputs.

\subsection{The Seventy Narrative-Spatial Features}
\label{sup:narrative_features}

Unassigned words are removed and the remaining time-aligned words are
ordered by onset. Adjacent entries with the same anchor form one visit,
spanning the first onset to the last offset. Intervening unassigned words
and long pauses do not split a visit; there is no elapsed-time cutoff. Consecutive visits define the narrative sequence
$v_i=(r_i,s_i,e_i,c_i)$, where $r_i$ is the anchor, $[s_i,e_i]$ its speech
interval, and $c_i$ the anchor's fixed image centroid. Define dwell duration
$d_i=e_i-s_i$, transition distance $a_i=\|c_{i+1}-c_i\|_2$, and nonnegative
transition pause $g_i=\max(0,s_{i+1}-e_i)$. Each centroid is the center of its fixed bounding box in reference-image
pixels, including the box associated with dish drying. Transition distances
therefore describe positions on the reference image.

\begin{table}[t]
\caption{The seventy narrative-spatial candidate features comprise $40+3+10+14+3$ descriptors. The forty-feature narrative-profile group is included within this total. The family-removal analysis evaluates all seventy descriptors together;
Appendix~\ref{sup:review_controls} examines specified smaller alternatives.}
\label{tab:supp_narrative_families}
\centering
\small
\begin{tabular}{@{}p{0.26\linewidth}rp{0.57\linewidth}@{}}
\toprule
Subfamily & Count & Represented property \\
\midrule
Narrative profile & 40 & Coverage, duration, path, graph structure, and mapping status \\
Reference deviation & 3 & Coverage, first-mention order, and dwell allocation relative to a healthy-control reference \\
Timing--space interaction & 10 & Relationships between pauses, dwell, and anchor transitions \\
Temporal segmentation & 14 & Changes between the first and second halves of the visit interval \\
Transition sequences & 3 & Repetition and entropy of anchor pairs and triples \\
\midrule
Total & 70 & \\
\bottomrule
\end{tabular}
\end{table}

\paragraph{Narrative profile: forty features.}
This subfamily contributes $3+21+6+7+3=40$ of the seventy narrative-spatial
features: coverage, timing, path, graph, and mapping status, respectively.
Three coverage features are the number of unique anchors, that number divided
by fifteen, and the number of unvisited critical anchors. The critical set is
boy, girl, woman, stool, cookie jar, cookies, sink, and water. Twenty-one timing
features comprise the mean, standard deviation, coefficient of variation,
maximum, and minimum visit duration; fifteen anchor-specific sums of duration;
and the onset of the first visit. Six path features are the transition count,
mean and standard deviation of transition distance, total path length,
unique-anchor count divided by transition count, and the fraction of visits
to an already visited anchor.

Seven graph features use a directed graph whose nodes are visited anchors and
whose edges are observed consecutive transitions. Transition counts are
retained as edge weights. The descriptors are node count, edge count, graph
density, maximum in-degree, maximum out-degree, entropy of normalized total
degrees, and the size of the largest strongly connected component. Finally,
three mapping-status features are failed grounding divided by eligible
content words, non-groundable words divided by all words, and directly
groundable words divided by all words. These last three can encode behavior
of the extraction procedure as well as properties of the description; they
are not independent measurements of cognitive impairment.

\paragraph{Reference deviation: three features.}
For a supplied healthy-control reference set, the reference anchors are those
mentioned by at least half its participants. First-mention ranks are averaged
across reference participants, assigning rank fifteen to an unmentioned
anchor. The reference dwell distribution is the average normalized
anchor-duration distribution. For participant anchor set $R$, reference set
$R_0$, first-mention ranks $q,q_0$, and dwell distributions $p,p_0$, the three
descriptors are
\begin{align}
\delta_{\mathrm{coverage}} &= 1-\frac{|R\cap R_0|}{|R_0|}, \\
\delta_{\mathrm{order}} &= 1-\tau(q,q_0), \\
\delta_{\mathrm{time}} &= \sum_{r=1}^{15}p_r\log\frac{p_r}{p_{0r}}.
\end{align}
Kendall's $\tau$ compares the participant's observed anchors with their
reference ranks. The reference has a rank for every anchor, including the
rank-fifteen penalties for absent mentions used when averaging across controls;
therefore the comparison uses all anchors observed in the target participant,
not only anchors meeting the reference coverage threshold. Fewer than two
observed anchors gives order deviation two; an undefined correlation uses zero. Dwell distributions use a
$10^{-10}$ additive floor followed by normalization, and an empty-duration
sample has a uniform distribution. The reference is a fitted,
label-dependent object: its participant set must be specified for every
evaluation fold rather than inferred from the descriptor formula.

\paragraph{Timing--space interactions: ten features.}
The descriptors are the mean, standard deviation, and maximum of $g_i$;
fraction of gaps exceeding one second; mean of $g_i a_i$; Pearson correlation
between gap and distance; Pearson correlation between a visit's dwell and
its following gap; transitions per second over the visit interval; mean gap
on recorded revisit transitions; and fraction of transitions for which both
preceding dwell exceeds the sample median and the following gap exceeds one
second. A transition from visit $i$ to $i+1$ contributes to the revisit-gap mean
when its target anchor occurs among visits $1,\ldots,i-1$. For example,
in boy--sink--boy, the gap on sink--boy is included. The current source is
excluded from this comparison. This defines the transition-level statistic
separately from the visit-level revisit fraction.
Undefined correlations and summaries of empty transition sets are assigned
zero.

\paragraph{Temporal segments: fourteen features.}
The interval from the first visit onset to the last visit offset is divided
at its temporal midpoint; a visit belongs to the first half when its own
midpoint is at or before that boundary. Each half contributes five values:
unique-anchor count, coverage divided by fifteen, transition count,
path efficiency (unique-anchor count divided by the maximum of transition
count and one), and entropy of visit-anchor frequencies. Four further descriptors are
first-minus-second differences in coverage, transition count, and path
efficiency, and second-minus-first change in entropy.

\paragraph{Transition sequences: three features.}
Contiguous anchor pairs and triples are extracted from the visit sequence.
The descriptors are entropy of empirical pair frequencies, entropy of triple
frequencies, and the fraction of distinct pair types occurring more than
once. Pair and triple frequencies are computed separately within each
participant's visit sequence.

The representation combines content coverage, ordering, and timing because
two narratives may mention the same anchors in different sequences or with
different timing. Appendix~\ref{sup:review_controls} evaluates a specified
six-feature compact alternative and a two-coordinate removal under the same
classifier and selection procedure. These comparisons assess the implemented
inventory; neither the feature definitions nor the observed intervals
establish that all seventy descriptors are necessary.

\subsection{Image--Text Relevance}
\label{sup:visual_features}

Ten image--text features use frozen Contrastive Language--Image Pretraining
(CLIP), ViT-L/14 \citep{radfordLearningTransferableVisual2021}, with the shared
Cookie Theft image and participant utterances. They are the mean, standard
deviation, minimum, and maximum whole-image--utterance similarity; its slope
against utterance order; fractions above $0.25$ and below $0.18$; the mean
maximum anchor--utterance similarity; entropy of mean anchor similarities;
and the fraction of anchors with maximum utterance similarity above $0.22$.
Image and text embeddings are unit-normalized before their dot products.
Each anchor has three separate representations: a text prototype for word
assignment, a pixel centroid for path distance, and a fixed image crop for
image--text similarity. The image encoder embeds each crop, while the
anchor mapper uses the corresponding text prototype.
For entropy, negative mean anchor similarities are set to zero before
normalizing to sum to one; a zero sum yields zero entropy. Utterances are truncated to
300 characters and then to 77 model tokens.
All participants share the same visual stimulus and fixed anchor crops.

\section{Likelihood, Context, and Missing Evidence}
\label{sup:likelihood}

\subsection{Target-Only Event Scoring}

The language model is the frozen \texttt{Qwen/Qwen2-7B} base causal model
\citep{qwen2}; no disease-specific fine-tuning is performed. Both model and
\texttt{Qwen2TokenizerFast} use the archived repository revision
\texttt{453ed1575b739b5b03ce3758b23befdb0967f40e}. The tokenizer's character
offsets locate scored target tokens. The archived inference configuration
uses bfloat16 and scaled dot-product attention.
Given a prompt
context $C$, target event text $x$, and its target-token index set $\mathcal{T}_e$, the score is
\begin{equation}
\mathcal{L}_C(\mathcal{T}_e)=\frac{1}{|\mathcal{T}_e|}\sum_{j\in\mathcal{T}_e}\log p_{\theta}(x_j\mid C,x_{<j}),
\end{equation}
in nats per token. Prompt headers and exemplars condition the prediction;
the average includes target tokens only. The same frozen model and scoring
operation are used for the two class contexts.

For each event, class-specific exemplar pools contain event text from
permitted training participants. Disease labels determine the pool outside
the prompt; no disease name or diagnostic label is serialized in the prompt.
Both classes use the following layout:
\begin{minipage}{\linewidth}
\begin{verbatim}
CONTEXT EXAMPLES
================
1. [Example event text 1]
...
K. [Example event text K]

NOW CONSIDER THE FOLLOWING UTTERANCE
====================================
[Target event text]
\end{verbatim}
\end{minipage}
Each exemplar occupies one numbered line. A blank line precedes the target
header and only a terminal newline follows the target text. Let
$n=\min(n_{\mathrm{AD}},n_{\mathrm{HC}})$ be the smaller eligible nonempty
class pool. Both classes use the same $K_{\mathrm{ctx}}$:
$K_{\mathrm{ctx}}=n$ when $n<10$, otherwise
$K_{\mathrm{ctx}}=\min(30,\lfloor0.70n\rfloor)$, sampled without replacement. For each
event and seed, two independently drawn context pairs produce two contrasts,
which are averaged. Each draw samples one context from each class; two draws
therefore give two AD/HC context pairs and two contrasts to average.
Fifty seeds vary
exemplar sampling and classifier randomness; the participant fold partition
has fixed random state $42$. The declared context limit is 4,096 tokens.
Per-instance prompt lengths were not retained in the historical logs;
approximately 930 tokens at the thirty-exemplar cap is a descriptive estimate,
not a verified maximum for every prompt.

\paragraph{Self-exclusion and fold exclusion are different requirements.}
Self-exclusion removes every span from the participant being scored from
both class pools, including when that participant belongs to a classifier's
training set. Fold exclusion additionally prevents any validation participant
from contributing exemplars to \emph{any} feature used to train or validate
the classifier in that fold. A single precomputed matrix whose rows were
individually scored out of fold does not necessarily satisfy this second
requirement after repartitioning: a training row's context may have contained
a participant now in validation. The classifier training and validation rows
must therefore use the same explicitly defined outer-training exemplar pool,
with additional per-target self-exclusion where applicable.

\subsection{The Twenty Likelihood-Stream Features}
\label{sup:likelihood_features}

Table~\ref{tab:likelihood_inventory} maps the same twenty columns to their
construction and experimental roles. The construction groups count
$4+9+6+1$; the targeted controls regroup them as twelve score-derived,
two coverage, and six surprisal columns. Reaggregation concerns the thirteen
class-context columns, including cloze coverage; the other seven are reused
because they do not depend on exemplar pools.

\begin{table}[htbp]
\caption{One inventory of the twenty likelihood-stream columns. ``Reaggregated'' means recomputed from existing class-context scoring records; ``reused'' means copied from saved features without new model inference. Positive-event counts depend on score signs.}
\label{tab:likelihood_inventory}
\centering\small
\begin{tabularx}{\linewidth}{@{}Xrll@{}}
\toprule
Descriptors & Count & Control group & Evaluation handling \\
\midrule
Cloze mean, minimum, positive count & 3 & Event scores & Reaggregated \\
Cloze defined-event coverage & 1 & Coverage & Reaggregated \\
Seven paired contrasts, mean, positive count & 9 & Event scores & Reaggregated \\
Scene mean, maximum, variance; vague, action, object surprisal & 6 & Surprisal & Reused \\
Unmatched-concept count & 1 & Coverage & Reused \\
\bottomrule
\end{tabularx}
\end{table}

\paragraph{Four cloze summaries.}
For events E1--E6, a fixed filler lexicon identifies positions in the retained
participant text. The event's hand-authored template defines this lexicon; the
template itself is not the scored target. The cloze contrast is
$\mathcal{L}_{C_{\mathrm{AD}}}(\mathcal{F}_e)-\mathcal{L}_{C_{\mathrm{HC}}}(\mathcal{F}_e)$, where $\mathcal{F}_e$ contains subword
positions whose original character substrings match normalized canonical
fillers. Selection compares each subword's character substring with the normalized
filler lexicon. It is not whole-word matching followed by inclusion of every
subword of that word. This differs from transcript-surprisal selection below. Its four aggregate descriptors are
the mean and minimum finite event contrast, the fraction of the six events
with a defined contrast, and the number of defined contrasts greater than
zero. E7 has no cloze template. No matched filler produces an undefined
contrast rather than a neutral zero.

\paragraph{Nine paired-likelihood features.}
Each of E1--E7 has the full-target contrast
$\mathcal{L}_{C_{\mathrm{AD}}}(\mathcal{T}_e)-\mathcal{L}_{C_{\mathrm{HC}}}(\mathcal{T}_e)$. Their mean and count of positive
finite contrasts add two summaries. An event with no retained text has an
undefined paired contrast. A positive contrast means that the frozen model
assigns greater average log likelihood under the AD exemplar context; it is
not an event-level clinical diagnosis or a causal contribution to the final
classifier score.

\paragraph{Six transcript-context surprisal summaries.}
The participant transcript is also scored under its own preceding textual
context. Fixed object, actor, action, and vague-term lexicons select token
positions in two steps: first match full words to these lexicons, then include
subwords whose character intervals lie wholly within the matched word intervals.
Unlike cloze selection, this step does not independently match each subword
surface to a filler lexicon. With surprisal
$u_j=-\log p_{\theta}(x_j\mid x_{<j})$, the descriptors are mean, maximum,
and variance over the union of object, actor, and action positions, plus mean
surprisal over vague, action, and object positions separately. These six
values use transcript context and scene-lexicon selection; they are not
exemplar-conditioned local event scores.
Transcripts longer than 256 tokens use windows of 256 with 64-token overlap
and stride 192. The first window emits its scored positions; later windows
emit positions after their 64-token overlap, so every scored transcript
token contributes once. The first transcript token has no preceding context
and its undefined score is excluded from finite-score summaries.

\paragraph{One unmatched-concept count.}
Canonical concept clusters specify lexical alternatives. The feature counts
clusters not recovered from scene-lexicon matches in the transcript. It therefore measures
unmatched canonical content under a particular normalization and lexicon.
It does not distinguish genuine participant omission from a false-negative
lexical match.

\subsection{Normalization and Missingness Indicators}
\label{sup:missingness}

Likelihood matching lowercases tokens, strips boundary punctuation, and
applies the first eligible suffix rule in the order \texttt{ies},
\texttt{ied}, \texttt{ing}, \texttt{ed}, \texttt{es}, \texttt{er}, and
\texttt{s}. This is the normalization used for the likelihood lexicons; anchor assignment
and event-field matching have their own rules. A word matched by one channel
is therefore not automatically a match in another. It has no irregular-word table or doubled-consonant recovery:
for example, \emph{running} becomes \emph{runn}. Consequently, a spoken
concept may fail a canonical match even when a human would recognize it.

For each of the twenty columns, a nonfinite value is replaced by a median
estimated from the fitting partition and accompanied by a binary flag. A
wholly missing fitting column uses zero as the numeric fallback. The flags
retain whether the original quantity was defined; all forty resulting
columns are eligible for selection. Other evidence families do not receive
these flags. Therefore, deleting the entire twenty-feature stream also
changes the availability of missingness signals. A comparison that deletes
the stream cannot attribute its effect solely to language-model likelihood
values without a separate missingness-only control.

\section{Classifier and Evaluation Boundaries}
\label{sup:evaluation}

\subsection{Fixed Classifier and Feature Selection}

The classifier specification is a support vector machine with radial basis
function kernel, $C=30$, $\gamma=0.001$, and $K_{\mathrm{feat}}=60$. The kernel is
$\exp(-\gamma\|z-z'\|_2^2)$ on standardized selected features. Feature ranking
averages two ordinal ranks: decreasing estimated mutual information with the
class label, and increasing two-sided Mann--Whitney test $p$-value. The sixty
smallest average ranks are retained. Both quantities depend on labels and
must be fitted using only the applicable training partition.

The support vector machine estimates probabilities by fitting a sigmoid to
its decision scores through internal five-fold fitting. This produces
probability estimates rather than raw distances from the decision boundary.
There is no additional shared calibration model after the classifier.
Classifier randomness includes this internal probability estimation.

Median estimates, supervised feature ranks, standardization, healthy-control
references, and exemplar pools each have a fitting scope. Generating
classifier out-of-fold probabilities after preprocessing all 108 training
participants is insufficient to call the entire procedure out of fold.
The fitting scope of each stage must be specified independently of whether
official test labels were used. The fixed classifier configuration is not
presented here as the outcome of an independently validated ranking of eight
backends.

\paragraph{Prior test exposure.}
Official-test model comparisons and feature-family analyses preceded the
corrective evaluation. The correction retained the classifier and its
parameters, but the same test participants had already been inspected;
the corrected results therefore remain exploratory. The development record
does not establish which design choices were influenced by test performance.

\subsection{Corrective Fitting Procedure}

\begin{table}[htbp]
\caption{Data available at each fitting stage. An outer validation fold is one fifth of the 108 official training participants; the official test participants never enter these training pools. ``Own fold'' is the outer fold containing the target participant.}
\label{tab:fitting_scopes}
\centering\small
\begin{tabularx}{\linewidth}{@{}L{0.19\linewidth}XX@{}}
\toprule
Stage / target row & Exemplars and healthy-control reference & Preprocessing and classifier \\
\midrule
Outer fitting rows & Other four folds; also exclude the target itself & Fit medians, ranks, scaler, and classifier on these rows \\
Outer validation rows & Same other four folds & Apply that outer fit; save held-out probabilities \\
Final training rows & Each target's own-fold-excluded version & Fit final preprocessing and classifier on all 108 rows \\
Official test rows & All 108 training participants & Apply the final fit \\
Internal probability folds & Keep the feature values already supplied to the classifier & Fit the sigmoid internally; do not rebuild references or upstream selection \\
\bottomrule
\end{tabularx}
\end{table}

The correction fixes the historical model and all family masks before
reanalysis; it does not search configurations against test accuracy.
Only the downstream fitting and the required summaries are recomputed:
class-context language-model scores already exist for the allowed outer-fold
pools. They are filtered and reaggregated, not generated again. Table~\ref{tab:fitting_scopes}
separates the outer folds from the classifier's internal probability folds.
For each of the fifty seeds and each of the five fixed outer training folds:
\begin{enumerate}
\item Reaggregate the four cloze and nine paired descriptors for every
training participant from persisted scores conditioned on the same held-out
fold. No participant
in that fold contributes an exemplar to any fitting or validation row. A
fitting target is additionally excluded from its own contexts.
\item Rebuild the healthy-control reference from the fitting partition only.
Exclude a fitting target from its own reference. Validation participants use
the full reference allowed by that fold.
\item Apply the fixed family mask. Fit likelihood medians and missingness
augmentation, supervised top-sixty selection, standardization, and the
support vector classifier using fitting rows only. Predict the held-out rows.
\end{enumerate}
Concatenating the five validation predictions gives one probability for each
of the 108 training participants. The final classifier uses all 108 rows,
with each row's own held-out-fold likelihood and healthy-control reference.
The 48 test rows use full-training contexts and the full-training control
reference. All final preprocessing quantities are fitted on the 108 training
rows. Thus, the final fitting and test rows use different reference-pool
sizes; the correction does not claim to eliminate this distributional
difference.

The support vector classifier's internal five-fold sigmoid fitting occurs
within its allowed fitting partition. These internal folds do not separately
refit the upstream feature selector or class-conditioned feature construction.
This probability mechanism is retained from the historical specification,
without claiming independently established calibration quality.

Thirteen of the twenty likelihood-stream columns are thus reaggregated from
the class-context scoring logs. The six transcript-context surprisal columns
and unmatched-concept count are reused from accepted feature tables, with
exact equality checked against the archived combined features. These seven
columns do not depend on the class exemplar pools. No new language-model
inference was performed for this correction; reaggregation does not constitute
an independent reproduction of token-level model scores.

\paragraph{Computational checks.}
Independent scripts checked saved participant splits, reference memberships,
selected-feature records, probabilities, thresholds, decisions, and reported
metrics. For the targeted controls in Appendix~\ref{sup:review_controls}, a
separate statistical script recomputed result rows and paired comparisons
from saved predictions and the original bootstrap indices. These checks did
not refit models, regenerate language-model scores or bootstrap draws,
independently rerun mutual-information ranking, or validate event semantics.
They establish computational consistency rather than independent replication,
external generalization, or human inspection benefit. Original scoring
records lack exemplar identifiers; exemplar isolation is supported by the
frozen generator and fold keys rather than a per-record membership audit.

\subsection{Metrics and Participant-Level Uncertainty}
\label{sup:stats}

For fifty fitted seeds, mean accuracy and mean area under the receiver
operating characteristic curve (AUC) average the fifty per-seed values, and
population standard deviation describes variation across those seeds. A per-seed maximum
is descriptive and is not an unbiased estimate of generalization. Aggregated
metrics use each participant's mean predicted probability across the fifty
seeds; they are different estimands from the mean of per-seed metrics.

The decision threshold maximizes Youden's statistic, sensitivity plus
specificity minus one \citep{Youden}, over the grid $0.10$ to $0.90$ with step
$0.01$, using the concatenated training validation probabilities; the smallest
threshold resolves ties. AUC does not depend on this
threshold. Threshold-dependent results require the verified prediction and
threshold provenance described above. The official test partition contains
48 participants, so one changed correct decision corresponds to approximately
2.08 percentage points of accuracy. Fifty seeds do not increase the number
of independent test participants.

Participant uncertainty uses 10,000 diagnosis-stratified bootstrap resamples
with random seed 20260925, paired across systems and holding thresholds fixed.
For mean accuracy, a participant contributes the fraction of fifty seed
decisions that were correct; aggregated accuracy instead uses one binary
correctness value, and aggregated AUC uses the mean probability. Percentile
limits at 2.5\% and 97.5\% form the intervals. Such intervals describe participant sampling
conditional on the fitted systems; they do not capture all uncertainty from
training-set selection or methodological development. Exact two-sided
McNemar tests compare discordant correctness decisions
\citep{McNemar_1947}. If there are no discordant decisions, the comparison is
reported as degenerate rather than used as evidence of equality.

\section{Corrected Participant-Level Statistics}
\label{sup:corrected_statistics}

The following tables report the verified corrective evaluation on the 48
official-test participants. Alzheimer's disease (AD) is the positive class;
healthy control (HC) is the negative class. All intervals use the
diagnosis-stratified participant bootstrap specified in
Section~\ref{sup:stats}, with predictions and thresholds held fixed.
They quantify participant sampling uncertainty conditional on the fitted
systems, rather than variation across random seeds.

\subsection{Uncertainty and Operating Points}

Table~\ref{tab:corrected_participant_intervals} gives point estimates and
95\% confidence intervals. Table~\ref{tab:corrected_operating_points} reports
the training-derived threshold and the corresponding sensitivity and
specificity after averaging probabilities across fifty seeds. Sensitivity
is the fraction of AD participants classified as AD; specificity is the
fraction of HC participants classified as HC.

\begin{table}[htbp]
\caption{Corrected point estimates with participant-bootstrap 95\% confidence
intervals underneath. Accuracy and area under the receiver operating
characteristic curve (AUC) are shown on a 0--100 scale. Mean accuracy
averages fifty seed-specific accuracies; aggregated metrics use each
participant's mean probability.}
\label{tab:corrected_participant_intervals}
\centering
\small
\begin{tabular}{@{}L{0.28\linewidth}C{0.20\linewidth}C{0.20\linewidth}C{0.20\linewidth}@{}}
\toprule
System & \shortstack{Mean\\accuracy} & \shortstack{Aggregated\\accuracy} & \shortstack{Aggregated\\AUC} \\
\midrule
Full & \shortstack{80.88\\$[70.75,89.88]$} & \shortstack{81.25\\$[68.75,91.67]$} & \shortstack{91.32\\$[82.47,97.57]$} \\
Without speech timing & \shortstack{83.25\\$[73.63,91.75]$} & \shortstack{85.42\\$[75.00,93.75]$} & \shortstack{92.19\\$[83.68,98.26]$} \\
Without acoustic & \shortstack{81.08\\$[70.83,90.17]$} & \shortstack{79.17\\$[66.67,89.58]$} & \shortstack{91.15\\$[82.12,97.40]$} \\
Without linguistic & \shortstack{80.88\\$[71.29,89.63]$} & \shortstack{81.25\\$[68.75,91.67]$} & \shortstack{90.97\\$[81.77,97.57]$} \\
Without image--text & \shortstack{81.25\\$[71.29,90.21]$} & \shortstack{79.17\\$[66.67,89.58]$} & \shortstack{91.84\\$[83.33,97.74]$} \\
Without narrative-spatial & \shortstack{78.21\\$[67.58,87.79]$} & \shortstack{81.25\\$[68.75,91.67]$} & \shortstack{87.67\\$[76.56,96.18]$} \\
Without likelihood stream & \shortstack{81.63\\$[71.46,90.75]$} & \shortstack{81.25\\$[70.83,91.67]$} & \shortstack{88.89\\$[78.30,97.05]$} \\
\bottomrule
\end{tabular}
\end{table}

\begin{table}[htbp]
\caption{Operating points for aggregated predictions. Thresholds are on the
0--1 probability scale; sensitivity and specificity are percentages.
Each system uses its own training-derived threshold.}
\label{tab:corrected_operating_points}
\centering
\small
\begin{tabular}{@{}L{0.42\linewidth}rrr@{}}
\toprule
System & Threshold & Sensitivity & Specificity \\
\midrule
Full & 0.56 & 75.00 & 87.50 \\
Without speech timing & 0.59 & 79.17 & 91.67 \\
Without acoustic & 0.62 & 70.83 & 87.50 \\
Without linguistic & 0.41 & 79.17 & 83.33 \\
Without image--text & 0.51 & 75.00 & 83.33 \\
Without narrative-spatial & 0.47 & 83.33 & 79.17 \\
Without likelihood stream & 0.45 & 87.50 & 75.00 \\
\bottomrule
\end{tabular}
\end{table}

\subsection{Paired Family-Removal Comparisons}

Table~\ref{tab:corrected_paired_differences} reports the full system minus
each family-removal system, so a positive difference favors the full
system. The resampled participant indices are shared by the two systems in
each comparison. Every displayed paired interval includes zero.
Table~\ref{tab:corrected_discordances} reports the complete discordant
decision counts: \emph{improved} means full correct and removal incorrect;
\emph{regressed} means full incorrect and removal correct.
The exact two-sided McNemar test applies to these aggregated decisions,
not to mean-over-seed correctness.

\begin{table}[htbp]
\caption{Full-minus-removal differences and paired participant-bootstrap
95\% confidence intervals underneath. All values are differences on the
0--100 metric scale: percentage points for accuracy and AUC points for
AUC. Positive values favor the full system.}
\label{tab:corrected_paired_differences}
\centering
\small
\begin{tabular}{@{}L{0.28\linewidth}C{0.20\linewidth}C{0.20\linewidth}C{0.20\linewidth}@{}}
\toprule
Removed family & \shortstack{Mean\\accuracy} & \shortstack{Aggregated\\accuracy} & \shortstack{Aggregated\\AUC} \\
\midrule
speech timing & \shortstack{-2.38\\$[-6.63,0.46]$} & \shortstack{-4.17\\$[-10.42,0.00]$} & \shortstack{-0.87\\$[-4.86,2.78]$} \\
acoustic & \shortstack{-0.21\\$[-1.54,1.25]$} & \shortstack{2.08\\$[0.00,6.25]$} & \shortstack{0.17\\$[-1.56,2.08]$} \\
linguistic & \shortstack{0.00\\$[-2.71,3.17]$} & \shortstack{0.00\\$[-6.25,6.25]$} & \shortstack{0.35\\$[-2.43,3.47]$} \\
image--text & \shortstack{-0.38\\$[-2.08,1.25]$} & \shortstack{2.08\\$[0.00,6.25]$} & \shortstack{-0.52\\$[-3.13,1.56]$} \\
narrative-spatial & \shortstack{2.67\\$[-3.67,9.00]$} & \shortstack{0.00\\$[-10.42,10.42]$} & \shortstack{3.65\\$[-2.78,10.59]$} \\
likelihood stream & \shortstack{-0.75\\$[-9.33,7.71]$} & \shortstack{0.00\\$[-12.50,12.50]$} & \shortstack{2.43\\$[-4.86,9.90]$} \\
\bottomrule
\end{tabular}
\end{table}

\begin{table}[!htbp]
\caption{Discordant aggregated decisions comparing the full system with
each family removal. Counts are participants; $p$ is the exact two-sided
McNemar test $p$-value.}
\label{tab:corrected_discordances}
\centering
\small
\begin{tabular}{@{}L{0.42\linewidth}rrr@{}}
\toprule
Removed family & Improved & Regressed & $p$ \\
\midrule
speech timing & 0 & 2 & 0.500 \\
acoustic & 1 & 0 & 1.000 \\
linguistic & 1 & 1 & 1.000 \\
image--text & 1 & 0 & 1.000 \\
narrative-spatial & 3 & 3 & 1.000 \\
likelihood stream & 5 & 5 & 1.000 \\
\bottomrule
\end{tabular}
\end{table}

\subsection{Every Changed Decision in the Likelihood-Stream Comparison}

Table~\ref{tab:corrected_likelihood_changes} lists all ten participants whose
aggregated predicted class differs between the full system and the system
without the likelihood stream. The full-system threshold is $0.56$ and the
without-likelihood threshold is $0.45$. The latter removal excludes both the
twenty likelihood-stream features and their twenty missingness indicators.

Five decisions improve: four HC participants and one AD participant.
Five regress: four AD participants and one HC participant. Both systems
therefore classify 39 of 48 participants correctly, although their
sensitivities and specificities differ. These are all changed decisions,
not a selection of successful examples. The comparison concerns two refitted
systems and their respective thresholds; it is not a causal attribution to
an individual event score.

\paragraph{Reading a changed decision.}
S197 illustrates why probability changes and decision changes must be read
together. This healthy-control participant's AD probability rises from
$0.4695$ without the likelihood stream to $0.5163$ with it, yet the
classification changes from AD to HC because the training-selected threshold
also rises from $0.45$ to $0.56$. The corrected decision therefore cannot be
described as the stream lowering the estimated disease probability. By
contrast, S168 has a true AD label and a full-system probability of $0.5184$,
just above S197's value, and becomes a false negative under the same $0.56$
threshold. Inspecting only the successful S197 case would hide this nearby
failure. The complete case table preserves both directions of change.

\begin{table}[!htbp]
\caption{All ten changed aggregated decisions. Probabilities are the mean
AD probability over fifty seeds, rounded to four decimals for display;
decisions use the unrounded values and thresholds $0.45$ (without likelihood)
and $0.56$ (full). Change is defined when moving from without likelihood to
the full system.}
\label{tab:corrected_likelihood_changes}
\label{tab:changed_likelihood_cases}
\centering
\small
\begin{tabular}{@{}lcrcrcl@{}}
\toprule
& & \multicolumn{2}{c}{Without likelihood} & \multicolumn{2}{c}{Full system} & \\
\cmidrule(lr){3-4}\cmidrule(lr){5-6}
Sample & True & $p_{\mathrm{AD}}$ & Decision & $p_{\mathrm{AD}}$ & Decision & Change \\
\midrule
S162 & AD & 0.6159 & AD & 0.2339 & HC & Regressed \\
S164 & AD & 0.3121 & HC & 0.7523 & AD & Improved \\
S168 & AD & 0.5966 & AD & 0.5184 & HC & Regressed \\
S169 & AD & 0.4828 & AD & 0.3436 & HC & Regressed \\
S170 & HC & 0.5071 & AD & 0.3320 & HC & Improved \\
S175 & HC & 0.5593 & AD & 0.3814 & HC & Improved \\
S177 & HC & 0.1066 & HC & 0.6675 & AD & Regressed \\
S180 & HC & 0.7724 & AD & 0.4042 & HC & Improved \\
S189 & AD & 0.6615 & AD & 0.4080 & HC & Regressed \\
S197 & HC & 0.4695 & AD & 0.5163 & HC & Improved \\
\bottomrule
\end{tabular}
\end{table}

\section{Published References and Historical Comparisons}
\label{sup:controls}

\subsection{Published Reference Results}
\label{sup:published_reference}

\paragraph{The three Xiao--Mistral scores.}
For a transcript, let $P_{\mathrm{AD}}$ and $P_{\mathrm{HC}}$ be its perplexities
under the separately class-adapted models, and let $d=P_{\mathrm{AD}}-P_{\mathrm{HC}}$.
Let $\mu_c$ and $\sigma_c$ be the training-class mean and standard deviation
of $d$. The reported scores of \citet{xiaoAlzheimersDementiaDetection2025} are
\begin{align}
\tilde D_{\log}&=\frac{\log P_{\mathrm{HC}}-\log P_{\mathrm{AD}}}{\log P_{\mathrm{HC}}},\\
\bar D&=|d-\mu_{\mathrm{HC}}|-|d-\mu_{\mathrm{AD}}|,\\
\bar D^*&=|d-(\mu_{\mathrm{HC}}-2\sigma_{\mathrm{HC}})|
-|d-(\mu_{\mathrm{AD}}+2\sigma_{\mathrm{AD}})|.
\end{align}
Thus the first normalizes a log-perplexity difference; the other two compare
distance to training-class means, with the starred version shifting those
means by two standard deviations. The historical reproduced accuracies use
test-derived equal-error operating points, not the current NarraSpace
training-selected threshold.

Table~\ref{tab:published_context} retains the original literature results separately from the harmonized-transcript reproductions in Table~\ref{tab:main_comparison}. The official participant split alone does not align transcript preprocessing, training data, model adaptation, aggregation, or threshold estimation. SpeechHGT repartitions all 156 participants, the transfer model trains on CognoMemory, and Xiao--Mistral reports accuracy at an equal-error operating point derived from test scores and labels. NarraSpace uses training-selected thresholds. These differences preclude a common-protocol ranking; the present results do not establish a performance advantage over the published methods.

\begin{table}[htbp]
\caption{Published results under their original evaluation protocols. This table separates reported point estimates from explicitly reported fifty-seed statistics. These literature values provide context; they are not a common-protocol ranking with Table~\ref{tab:main_comparison}. A dash denotes an unreported metric. All values are percentages.}
\label{tab:published_context}
\centering
\small
\begin{tabularx}{\linewidth}{@{}X L{0.29\linewidth}rr@{}}
\toprule
Method & Protocol / reported summary & Accuracy & AUC \\
\midrule
\multicolumn{4}{@{}l}{\textbf{Published point estimates}} \\
\midrule
Official linguistic baseline \citep{luzAlzheimersDementiaRecognition2020} & Official 108/48 split & 75.00 & --- \\
SpeechHGT \citep{abidSpeechHGTMultimodalHypergraph2025} & New 70/30 split of all 156 & 86.32 & --- \\
Bidirectional long short-term memory transfer \citep{paharAutomaticDetectionEarly2025} & CognoMemory training; official 48 external test & 85.00 & 84.00 \\
TITAN \citep{liDetectingNeurocognitiveDisorders2025} & Official 108/48; last-five-epoch average & 89.58 & 94.10 \\
Pause-augmented ERNIE3p \citep{yuanPausesDetectionAlzheimers2021} & Official 108/48; 35-model majority vote & 89.60 & --- \\
Picture relevance \citep{zhuEvaluatingPictureDescription2023} & Official 108/48; reported best configuration & 89.58 & --- \\
\bottomrule
\end{tabularx}

\medskip
\begingroup
\setlength{\tabcolsep}{2.4pt}
\begin{tabular}{@{}L{0.25\linewidth}rrrrrr@{}}
\toprule
\multicolumn{7}{@{}l}{\textbf{Published Xiao--Mistral results \citep{xiaoAlzheimersDementiaDetection2025}}} \\
\multicolumn{7}{@{}l}{\emph{Official 108/48 split; fifty seeds; accuracy at the equal-error operating point}} \\
\midrule
& \multicolumn{3}{c}{Accuracy (\%)} & \multicolumn{3}{c}{AUC (\%)} \\
\cmidrule(lr){2-4}\cmidrule(lr){5-7}
Published score & Mean $\pm$ std. & Best & Aggregated & Mean $\pm$ std. & Best & Aggregated \\
\midrule
$\tilde{D}_{\log}$ & $86.37\pm3.61$ & 93.75 & 91.67 & $92.71\pm1.60$ & 95.31 & 92.71 \\
$\bar{D}$ & $85.96\pm3.33$ & 93.75 & 89.58 & $93.91\pm1.20$ & 97.05 & 94.97 \\
$\bar{D}^{*}$ & $85.96\pm3.33$ & 93.75 & 87.50 & $93.65\pm1.46$ & 97.05 & 89.76 \\
\bottomrule
\end{tabular}
\endgroup
\end{table}

\subsection{Scope of the Historical Backend Comparison}
\label{sup:restored_historical_comparison}

Table~\ref{tab:main_comparison}B preserves all three Xiao--Mistral reproductions and all eight NarraSpace backends with their complete six-metric summaries. Its NarraSpace rows share the 236-feature raw representation, with backend-specific historical feature selection. They retain the earlier fitting and feature-construction limitations and have not been rerun under the corrected partition-isolated procedure. In particular, the historical support vector machine has mean accuracy $80.25\%$, whereas the corrected primary result is $80.88\%$. The shared parser aligns transcript reconstruction; backbones, adaptation, feature dimensions, and threshold estimation remain specific to each method. Best-seed values describe observed maxima, and the two best-seed metrics can come from different seeds. Each row retains one concrete configuration.

For aggregated Xiao--Mistral results, AD-model and HC-model perplexities
are averaged separately across fifty runs for each participant before
applying the score definitions in Appendix~\ref{sup:published_reference}.
Accuracy uses a test-derived equal-error-rate threshold. NarraSpace instead
averages the fifty classifier probabilities for each participant, then
applies a threshold selected from training predictions. Aggregation thus
occurs before score construction for Xiao--Mistral and after classifier
probability estimation for NarraSpace. Appendix~\ref{sup:evaluation}
specifies the primary system's fitting procedure; block B retains the
historical fitting scope described above.

\subsection{Historical Whole-Transcript and Permutation Controls}

The controls in this subsection belong to historical evaluation pipelines.
The whole-transcript inputs do not have certified fold-specific upstream
model provenance for the corrected downstream threshold procedure. Their
numbers are therefore retained separately in Table~\ref{tab:historical_controls},
not inserted into the corrected block of the main table or treated as a fair
comparison with newly corrected predictions.

The six replacement inputs are HC-model perplexity, AD-model perplexity,
their AD-minus-HC difference, their HC/AD ratio, the HC-minus-AD log difference,
and that log difference divided by log HC perplexity. They are not six
versions of the three classification scores in Table~\ref{tab:main_comparison}.
Replacing the twenty likelihood-stream descriptors with these six
Xiao--Mistral whole-transcript descriptors
\citep{xiaoAlzheimersDementiaDetection2025} simultaneously changes backbone,
adaptation, scoring granularity, and dimension. The resulting raw pool has
222 columns, or 228 after six missingness indicators. This comparison changes both
the model and the scoring unit.
A matched locality experiment would hold the language model, fitting data,
exemplar construction, classifier, and missingness treatment fixed while
varying the scoring unit.

The fixed cyclic control applies
E1$\leftarrow$E4, E2$\leftarrow$E5, E3$\leftarrow$E6,
E4$\leftarrow$E7, E5$\leftarrow$E1, E6$\leftarrow$E2,
and E7$\leftarrow$E3 to the seven paired columns in both training and test
data, followed by refitting. A shared column permutation preserves their
values and much of the feature-space geometry. It is therefore a
permutation-invariance check of the fitted feature representation. Correct event names remain necessary for a semantically
accurate evidence report.

\begin{table}[htbp]
\caption{Historical descriptive results under the previous evaluation pipelines,
before the partition-isolation correction. These numbers are retained for
transparency and are not corrected head-to-head comparisons. Values are
percentages; the first two columns are means over fifty seeds.}
\label{tab:historical_controls}
\centering
\small
\begin{tabular}{@{}lrrrr@{}}
\toprule
 & \multicolumn{2}{c}{Fifty-seed mean} & \multicolumn{2}{c}{Aggregated probability} \\
\cmidrule(lr){2-3}\cmidrule(lr){4-5}
Historical configuration & Accuracy & AUC & Accuracy & AUC \\
\midrule
NarraSpace full representation & 80.25 & 90.70 & 81.25 & 91.32 \\
Whole-transcript replacement & 81.38 & 93.25 & 87.50 & 94.79 \\
Consistent event-column permutation & 80.71 & 90.94 & 81.25 & 91.32 \\
\midrule
Xiao--Mistral, $\tilde{D}_{\log}$ & 80.92 & 88.62 & 79.17 & 89.41 \\
Xiao--Mistral, $\bar{D}$ & 80.67 & 90.03 & 83.33 & 91.84 \\
Xiao--Mistral, $\bar{D}^{*}$ & 80.50 & 89.92 & 83.33 & 87.33 \\
\bottomrule
\end{tabular}
\end{table}

The whole-transcript replacement exceeded the historical full system on the
aggregated metrics. The result is specific to that cross-model comparison:
the reproduced Xiao--Mistral rows share transcript processing while retaining
different backbones, adaptation, feature dimensions, and fitting protocols. A matched-backbone locality
comparison remains unresolved.

The reproduced Xiao--Mistral accuracies use an equal-error operating point
estimated from the test scores and labels, rather than a threshold locked
from training predictions. This additional evaluation difference prevents
interpreting those accuracies as same-protocol comparisons with either the
historical or corrected NarraSpace classifier.

\section{Historical Transcript-Construction Comparison}
\label{sup:historical_transcripts}

The annotation-stripped reconstruction retains participant lines and removes
bracketed annotations, marked fillers, parenthetical content, and punctuation
other than periods and apostrophes, with a fixed phonetic-symbol substitution
list. It strips angle brackets but retains their enclosed words. The shared
NarraSpace parser instead removes the entire retraced angle-bracketed span
and its words, and preserves retained spoken forms as described in
Appendix~\ref{sup:transcript}. These are distinct reconstruction procedures,
with different retained spoken content.

Earlier whole-transcript reproductions used these distinct transcript
constructions and test-derived equal-error operating thresholds. Their
accuracies and historical NarraSpace reference do not establish a performance
difference under the corrected procedure, nor do they isolate transcript
construction as the sole explanation for differences from published results.

\section{Precision Audit of Accepted Anchor Mappings}
\label{sup:mapping_audit}

The mapping rules, prototypes, and acceptance thresholds were fixed before
the formal audit. A single annotator reviewed accepted phrase-rule and
prototype-based assignments from eighty participants. Sampling began with
twenty participants: seven training healthy controls, seven training
participants with Alzheimer's disease, three official-test healthy controls,
and three official-test participants with Alzheimer's disease. Fixed-seed
within-stratum top-up sampling was used to exceed the predefined minimum of
300 reviewed assignments. This yielded 404 judgments: 388 correct, sixteen
incorrect, and no unsure or unfilled judgments. Audit labels did not revise
the mapping rules or classifier.

Precision was 96.04\% overall, 100\% for phrase-rule assignments, and
92.45\% for prototype assignments. All sixteen errors occurred in the
prototype tier. These estimates apply only to the accepted phrase-rule and
prototype assignments sampled by the audit. They do not measure deterministic
lexical mapping, word-level timing, event assignment, actor--action binding,
or the accuracy of interpreting missing evidence. Unassigned candidates were
not annotated, so there is no false-negative denominator for recall or an
F1 score. One annotator provides no inter-annotator agreement estimate.

\begin{table}[t]
\caption{Per-anchor precision for the single-annotator audit of accepted
phrase-rule and prototype assignments. The cookies and curtain rows have only
one and four reviewed assignments; their percentages are not stable
anchor-specific estimates.}
\label{tab:mapping_audit_anchors}
\centering
\small
\begin{tabular}{@{}lrr@{}}
\toprule
Scene anchor & Correct / reviewed & Precision (\%) \\
\midrule
Boy & 33 / 33 & 100.00 \\
Girl & 43 / 44 & 97.73 \\
Woman & 14 / 15 & 93.33 \\
Stool & 25 / 26 & 96.15 \\
Cookie jar & 68 / 68 & 100.00 \\
Cookies & 0 / 1 & 0.00 \\
Sink & 12 / 13 & 92.31 \\
Water & 62 / 62 & 100.00 \\
Dishes & 42 / 45 & 93.33 \\
Curtain & 3 / 4 & 75.00 \\
Window & 20 / 24 & 83.33 \\
Counter & 10 / 10 & 100.00 \\
Floor & 31 / 34 & 91.18 \\
Garden & 9 / 9 & 100.00 \\
Dish-drying action & 16 / 16 & 100.00 \\
\bottomrule
\end{tabular}
\end{table}

Accepted-map precision describes the reviewed assignments, rather than the
fraction of all clinically relevant content recovered. A correctly spoken
paraphrase can remain unassigned and appear as an empty field in a downstream
report. Report labels therefore denote \emph{undetected evidence},
not verified participant omissions. Rejected candidates and event assignment
are outside this audit's sampling frame, so these results do not estimate
event-level recall or semantic accuracy.

\section{Evidence Reports and Reproducibility Scope}
\label{sup:cases}

Records link text, speech intervals, anchors, lexical hits, and likelihoods.
Scoring logs identify class context, fold, seed, and independent draw;
exact exemplar identifiers and serialized prompts require reconstruction
from frozen pools and randomization rules. Source links expose feature
construction through its recorded inputs and extraction choices. An AD-leaning
contrast can coexist with an HC prediction because the classifier combines
other selected features. Clinician usefulness and explanation faithfulness
have not been evaluated.

Table~\ref{tab:changed_likelihood_cases} enumerates all ten changed decisions.
Figure~\ref{fig:traceable_cases} shows the source records for two
participants classified correctly by both systems. Case records
use dataset sample codes rather than names.

\subsection{Implementation Availability}
\label{sup:code_scope}

The manuscript source archive contains the paper and figures, not the full
experimental pipeline or fitted artifacts. Reproducing participant predictions
requires authorized benchmark data, the specified model snapshots, transcript
parsing and alignment, feature extractors, split-aware orchestration, and
fitted preprocessing and classifiers. Feature definitions and evaluation
procedures are specified in Appendices~\ref{sup:impl}--\ref{sup:evaluation}.

Corrected classification used Python 3.10.12, NumPy 1.24.4, pandas 2.2.1,
SciPy 1.13.1, and scikit-learn 1.5.0. Computational checks and their limits
are summarized in Appendix~\ref{sup:evaluation}.

\section{Targeted Representation Controls After Test Exposure}
\label{sup:review_controls}

\subsection{Fixed Questions and Feature Pools}

We specified a finite set of controls after the corrective evaluation and
prior exposure to the official test set. The specification preceded this
control batch, not initial test access. All seven rows, including the full
reference, and all eight paired comparisons are reported. No result selects
a replacement primary system or changes its parameters.

The controls reuse the verified corrective feature caches, participant order,
five training folds, fifty seeds, and fitting procedure. Every pool uses the
same support vector classifier, $C=30$, $\gamma=0.001$, and sixty selected
inputs. Imputation, supervised ranking, standardization, classifier fitting,
and training-derived threshold selection are repeated within the permitted
partitions. The final training/test reference-pool asymmetry and internal
probability-fitting scope in Appendix~\ref{sup:evaluation} remain unchanged.
All twenty original likelihood missingness flags are constructed before any
numeric mask and retained as candidates in every row. Thus, these controls
differ from removing the entire likelihood family and its flags.

The controls regroup the twenty likelihood-stream columns as two coverage,
six transcript-context surprisal, and twelve event-score descriptors
(Table~\ref{tab:likelihood_inventory}). Positive-event counts belong to the
score group because they depend on likelihood signs. This grouping isolates
feature types; the thirteen reaggregated versus seven reused columns instead
describe dependence on training exemplars.
The three coverage-based rows retain all 216 inherited descriptors and the
indicated likelihood groups, together with all flags.

The compact spatial row replaces all seventy narrative-spatial descriptors
with six specified coordinates: unique-anchor count, total transitions,
revisit rate, mean transition distance, bigram entropy, and first-mention-order
deviation from the healthy-control reference. It retains the other 146 base
descriptors, all twenty likelihood numbers, and all flags. This is one
specified six-feature alternative evaluated with the same fitting procedure.
The two-coordinate removal drops graph node count and normalized anchor
coverage from the full pool. It retains narrative and linguistic anchor
counts, so it does not eliminate all redundant coordinates. A separate
mapping-quality control removes the failed-grounding, non-groundable-word,
and directly-groundable-word ratios defined in Appendix~\ref{sup:narrative_features}.

\begin{table}[t]
\caption{All targeted representation controls. Each of the three coverage-based rows retains the 216 base and narrative-spatial features in addition to the named groups. Candidate dimensions include
twenty missingness flags; every row selects sixty inputs. Performance values
are percentages, with population standard deviations over fifty seeds.
Thresholds are selected from training validation probabilities. These
post-exposure controls do not replace the fixed primary system.}
\label{tab:review_control_results}
\centering
\small
\begin{tabularx}{\linewidth}{@{}Xrcccc@{}}
\toprule
& & \multicolumn{2}{c}{Fifty-seed mean} & \multicolumn{2}{c}{Aggregated} \\
\cmidrule(lr){3-4}\cmidrule(lr){5-6}
Pool & Candidates & Accuracy & AUC & Accuracy & AUC \\
\midrule
Full reference & 256 & $80.88\pm2.56$ & $90.65\pm1.32$ & 81.25 & 91.32 \\
Coverage + flags & 238 & $79.33\pm0.57$ & $89.93\pm0.00$ & 79.17 & 89.93 \\
Coverage + surprisal + flags & 244 & $80.17\pm1.73$ & $92.01\pm0.00$ & 79.17 & 92.01 \\
Coverage + event scores + flags & 250 & $79.88\pm2.91$ & $89.73\pm1.81$ & 81.25 & 90.63 \\
Compact spatial & 192 & $77.96\pm3.78$ & $87.92\pm1.47$ & 83.33 & 89.06 \\
Two-coordinate removal & 254 & $80.96\pm2.64$ & $90.74\pm1.52$ & 81.25 & 92.19 \\
Mapping quality removed & 253 & $80.29\pm2.71$ & $90.60\pm1.33$ & 79.17 & 91.67 \\
\bottomrule
\end{tabularx}
\smallskip
\begin{tabularx}{\linewidth}{@{}Xccc@{}}
\toprule
Pool & Threshold & Sensitivity (\%) & Specificity (\%) \\
\midrule
Full reference & 0.56 & 75.00 & 87.50 \\
Coverage + flags & 0.40 & 87.50 & 70.83 \\
Coverage + surprisal + flags & 0.43 & 87.50 & 70.83 \\
Coverage + event scores + flags & 0.51 & 75.00 & 87.50 \\
Compact spatial & 0.46 & 87.50 & 79.17 \\
Two-coordinate removal & 0.42 & 79.17 & 83.33 \\
Mapping quality removed & 0.52 & 75.00 & 83.33 \\
\bottomrule
\end{tabularx}
\end{table}

\subsection{Paired Comparisons and Their Interpretation}

Table~\ref{tab:review_control_results} gives all point estimates and operating
points. Table~\ref{tab:review_control_paired_intervals} reports full minus each
of the six alternatives, plus the addition of surprisal or event scores to
coverage and flags. Intervals use the same 10,000 stratified participant
index draws as the corrective analysis, recomputing differences from the new
predictions. Fits and thresholds are held fixed within resampling.
The intervals are descriptive and unadjusted for these eight comparisons;
they do not capture repeated-test-exposure or development uncertainty.

Holding coverage, surprisal, and flags available, adding event-score numbers
changes mean accuracy by $0.71$ points, with interval $[-6.42,7.92]$, while
aggregated AUC changes by $-0.69$ points. The full-versus-compact comparison
also depends on the metric: full mean accuracy is higher by $2.92$ points,
but compact aggregated accuracy is $83.33\%$ versus $81.25\%$. Two-coordinate
removal has mean accuracy $80.96\%$ versus $80.88\%$ for full. None of the
eight mean-accuracy intervals excludes zero, and all exact McNemar tests
give $p=1.00$ (Table~\ref{tab:review_control_discordance}). These results
neither establish the necessity of the full design nor show that the compared
decision rules are equivalent. Each contrast includes refitted selection,
classifier, and threshold; it is not an individual-feature causal effect.

\begin{table}[t]
\caption{All eight paired representation-control contrasts, in percentage points.
Each cell gives left-minus-right difference and its unadjusted participant
bootstrap 95\% interval. ``Coverage,'' ``surprisal,'' and ``scores'' refer to
the numeric groups in Table~\ref{tab:likelihood_inventory}. Coverage-based pools also retain all 216 base and narrative-spatial features; every pool retains the twenty flags.}
\label{tab:review_control_paired_intervals}
\centering
\small
\begin{tabular}{@{}L{0.22\linewidth}C{0.23\linewidth}C{0.23\linewidth}C{0.23\linewidth}@{}}
\toprule
Left minus right & Mean accuracy & Aggregated accuracy & Aggregated AUC \\
\midrule
Full $-$ coverage & $1.54$ & $2.08$ & $1.39$ \\
 & $[-8.62,11.67]$ & $[-10.42,14.58]$ & $[-4.86,7.64]$ \\
\addlinespace
Full $-$ (coverage + surprisal) & $0.71$ & $2.08$ & $-0.69$ \\
 & $[-6.42,7.92]$ & $[-8.33,12.50]$ & $[-6.94,4.69]$ \\
\addlinespace
Full $-$ (coverage + scores) & $1.00$ & $0.00$ & $0.69$ \\
 & $[-4.75,6.75]$ & $[-8.33,8.33]$ & $[-4.17,5.38]$ \\
\addlinespace
Full $-$ compact spatial & $2.92$ & $-2.08$ & $2.26$ \\
 & $[-3.67,10.00]$ & $[-14.58,10.42]$ & $[-3.47,8.68]$ \\
\addlinespace
Full $-$ two-coordinate removal & $-0.08$ & $0.00$ & $-0.87$ \\
 & $[-5.71,5.79]$ & $[-10.42,10.42]$ & $[-5.73,3.47]$ \\
\addlinespace
Full $-$ quality removed & $0.58$ & $2.08$ & $-0.35$ \\
 & $[-0.75,2.00]$ & $[0.00,6.25]$ & $[-2.95,1.91]$ \\
\addlinespace
Coverage + surprisal $-$ coverage & $0.83$ & $0.00$ & $2.08$ \\
 & $[-5.79,7.58]$ & $[-8.33,8.33]$ & $[-2.43,7.64]$ \\
\addlinespace
Coverage + scores $-$ coverage & $0.54$ & $2.08$ & $0.69$ \\
 & $[-7.33,8.17]$ & $[-8.33,12.50]$ & $[-5.03,6.25]$ \\
\bottomrule
\end{tabular}
\end{table}

\begin{table}[t]
\caption{Hard-decision discordance for the same eight contrasts. Improved
means left correct and right incorrect; regressed means the reverse. Exact
two-sided McNemar tests concern aggregated decisions, not fractional
mean-over-seed correctness. No comparison has zero discordance.}
\label{tab:review_control_discordance}
\centering
\small
\begin{tabularx}{\linewidth}{@{}Xrrr@{}}
\toprule
Left minus right & Improved & Regressed & $p$ \\
\midrule
Full $-$ coverage & 6 & 5 & 1.00 \\
Full $-$ (coverage + surprisal) & 4 & 3 & 1.00 \\
Full $-$ (coverage + scores) & 2 & 2 & 1.00 \\
Full $-$ compact spatial & 4 & 5 & 1.00 \\
Full $-$ two-coordinate removal & 3 & 3 & 1.00 \\
Full $-$ quality removed & 1 & 0 & 1.00 \\
Coverage + surprisal $-$ coverage & 2 & 2 & 1.00 \\
Coverage + scores $-$ coverage & 4 & 3 & 1.00 \\
\bottomrule
\end{tabularx}
\end{table}

\subsection{Selection, Redundant Coordinates, and Retrieval Behavior}

Graph node count and linguistic anchor count each exactly equal narrative
anchor count for all 108 training and 48 test participants. Normalized anchor
coverage equals that count divided by fifteen and rounded to four decimals;
its maximum absolute difference from the unrounded ratio is
$3.33\times10^{-5}$. All four coordinates were selected in all fifty final
fits and all 250 outer-fold fits of the full reference. The two-coordinate
removal leaves two equal coordinates, so it tests that specified removal
while retaining the other equal pair.

Table~\ref{tab:selection_frequency} names the selection counts for the full
reference. None of the twenty missingness flags is selected in the fifty
final fits. Across 250 outer-fold fits, the E7 flag is selected fifty times
and the other flags never. The three mapping-quality ratios are absent from
all final fits, but two occur in outer-fold fits. Their removal therefore
need not leave the refitted procedure identical. Frequencies summarize
feature use across these correlated fits.

\begin{table}[htbp]
\caption{Full-reference feature-selection counts. Counts for grouped rows apply to each named feature separately. A dash denotes a count not tabulated here, not zero. Event names follow Table~\ref{tab:supp_events}.}
\label{tab:selection_frequency}
\centering\small
\begin{tabularx}{\linewidth}{@{}Xrr@{}}
\toprule
Feature & Final fits / 50 & Outer-fold fits / 250 \\
\midrule
Cloze mean & 49 & --- \\
Cloze minimum & 47 & --- \\
Cloze coverage & 0 & --- \\
Cloze positive-event count & 2 & --- \\
E1: cookie reaching & 35 & --- \\
E2: stool instability & 8 & --- \\
E3: girl observing & 31 & --- \\
E4: dishwashing; E5: overflow; E6: unawareness & 50 & --- \\
E7: kitchen context & 39 & --- \\
Paired mean; paired positive-event count & 50 & --- \\
Scene maximum surprisal; unmatched-concept count & 50 & --- \\
Other five surprisal descriptors & 0 & --- \\
E7 missingness flag & 0 & 50 \\
Other nineteen missingness flags & 0 & 0 \\
Failed-grounding ratio & 0 & 0 \\
Non-groundable-word ratio & 0 & 4 \\
Directly-groundable-word ratio & 0 & 50 \\
\bottomrule
\end{tabularx}
\end{table}

Table~\ref{tab:review_control_diagnostics} describes retrieval records with
participants as the group-summary unit and separate training/test strata.
The shared-utterance fraction is the fraction of a participant's unique
retrieved utterance identifiers assigned to more than one event. Group means
are approximately $79$--$82\%$; this is permitted by the retrieval rule and
describes overlap in the retrieved content. Record-completeness counts use 3,528
event--utterance records, where one utterance can occur under several events.
Every record contains nonempty text, a numeric valid interval, and complete
hit-field containers; 2,728 have a slot hit and 3,106 a nonempty region hit.
Field-container presence does not imply a positive hit. Numeric interval
validity does not establish word-timestamp accuracy.

\begin{table}[H]
\caption{Mechanical source descriptions, separated by official split and
diagnosis. Participant-level entries are group means: filled fields out of 35, nonempty events out of seven, and distinct retrieved utterances. Shared utterances are
those retrieved for multiple events, divided by unique retrieved utterances
within each participant, then averaged. Record counts describe field
availability; they are not semantic or timing accuracy estimates.}
\label{tab:review_control_diagnostics}
\centering
\small
\begin{tabularx}{\linewidth}{@{}Xrcccc@{}}
\toprule
Group & $n$ & Filled fields & Nonempty events & Unique utterances & Shared (\%) \\
\midrule
Training HC & 54 & 19.26 & 6.81 & 9.57 & 80.12 \\
Training AD & 54 & 14.65 & 6.20 & 8.44 & 79.08 \\
Test HC & 24 & 20.75 & 6.88 & 9.63 & 81.59 \\
Test AD & 24 & 15.42 & 6.71 & 7.54 & 81.11 \\
\bottomrule
\end{tabularx}
\smallskip
\begin{tabularx}{\linewidth}{@{}Xr@{}}
\toprule
Event--utterance record property & Count / 3,528 \\
\midrule
Nonempty text & 3,528 \\
Numeric valid time interval & 3,528 \\
All slot-hit fields present & 3,528 \\
Region-hit field present & 3,528 \\
At least one slot hit & 2,728 \\
Nonempty region hits & 3,106 \\
\bottomrule
\end{tabularx}
\end{table}

These summaries use the archived event cache, whose lineage was checked
against scoring inputs and the earlier training-only records. Those checks
do not independently audit the target text of every language-model call.
Group differences can reflect vocabulary, description length, or mapping
behavior. Without independent annotation, these records do not estimate
semantic precision, recall, false negatives, participant omissions, or
inter-annotator agreement.

\end{document}

%% file: authors.tex
\author{%
Qingming Li\thanks{Equal contribution.} \quad
Ruiyan Sun\footnotemark[1] \quad
Satoshi Nakamura\thanks{Corresponding author: \href{mailto:snakamura@cuhk.edu.cn}{snakamura@cuhk.edu.cn}}\\[0.5ex]
\normalfont\small The Chinese University of Hong Kong Shenzhen
}